%% file: main-5785-Adlakha.tex
\pdfoutput=1

\documentclass[11pt,a4paper]{article}
\usepackage{times,latexsym}
\usepackage{url}
\usepackage[T1]{fontenc}

\usepackage[acceptedWithA]{tacl2021v1}

\usepackage{times}
\usepackage{latexsym}
\usepackage{booktabs}
\usepackage{multirow, makecell}
\usepackage{graphicx}
\usepackage{enumitem}
\usepackage[colorinlistoftodos]{todonotes}

\usepackage{xcolor,colortbl}

\usepackage[T1]{fontenc}

\usepackage[utf8]{inputenc}

\usepackage{microtype}

\usepackage{inconsolata}

\usepackage{amssymb}
\usepackage{amsmath}
\usepackage[capitalise,nameinlink,noabbrev]{cleveref}
\usepackage{xcolor}
\usepackage{mdframed}
\usepackage{hyperref}
\usepackage{natbib}
\usepackage{placeins}
\usepackage[labelfont=bf]{caption}

\usepackage{array}
\newcolumntype{x}[1]{>{\centering\arraybackslash\hspace{0pt}}p{#1}}
\newcolumntype{M}[1]{>{\centering\arraybackslash}m{#1}}
\usepackage{tabularx}

\newcommand{\ignore}[1]{}

\newcommand{\checkspace}[1]{}

\newcommand{\revision}[1]{{#1}}

\title{Evaluating Correctness and Faithfulness of Instruction-Following Models for Question Answering}

\author{
\textbf{Vaibhav Adlakha$^{1,2}$ \hspace{7mm} Parishad BehnamGhader$^{1,*}$ \hspace{7mm} Xing Han Lu$^{1,*}$}\\
\textbf{Nicholas Meade$^{1,*}$ \hspace{7mm} Siva Reddy$^{1,2,3}$} \vspace{2mm} \\
$^{1}$Mila, McGill University \hspace{4mm} $^{2}$ServiceNow Research \hspace{4mm} $^{3}$Facebook CIFAR AI Chair
\vspace{2mm} \\
\texttt{\{firstname.lastname\}@mila.quebec} \\
}

\begin{document}
\maketitle
\renewcommand*{\thefootnote}{*}
\footnotetext[0]{Core contributor.}
\renewcommand*{\thefootnote}{\arabic{footnote}}
\setcounter{footnote}{0}
\begin{abstract} 
Instruction-following models are attractive alternatives to fine-tuned approaches for question answering (QA). By simply prepending relevant documents and an instruction to their input, these models can be adapted to various information domains and tasks without additional training. However, these models tend to produce verbose responses with supplementary information, which makes traditional QA metrics like exact match (EM) and F1 unreliable for accurately quantifying model performance.
In this work, we evaluate instruction-following models along two fronts: 1) how well they satisfy user's information need (correctness), and 2) whether they disseminate information supported by the provided knowledge (faithfulness). Guided by human evaluation and analysis, we highlight the shortcomings of traditional metrics for both correctness and faithfulness and propose simple token-overlap metrics that correlate highly with human judgments. Our analysis reveals that for correctness, instruction-following models perform comparably to models specifically fine-tuned for that task. However, they struggle to accurately judge the relevance of the provided knowledge and often hallucinate in their responses. We hope our work encourages more holistic evaluation of instruction-following models for QA.
Our code and human annotation data is available at \url{https://github.com/McGill-NLP/instruct-qa}.
\end{abstract}

\input{Sections/introduction}

\input{Sections/related_work}
\input{Sections/setup}
\input{Sections/correctness}

\input{Sections/groundedness}


\input{Sections/conclusion}
\input{Sections/acknowledgements}


\appendix
\input{Sections/appendix_experimental_details.tex}
\input{Sections/appendix_impact_of_retriever}
\input{Sections/appendix_real_world_holistic_eval}

\bibliographystyle{acl_natbib}
\bibliography{references}

\end{document}

%% file: Sections/introduction.tex
\section{Introduction}
\label{sec:introduction}

\begin{figure}[t!]
    \centering
    \includegraphics[width=\linewidth]{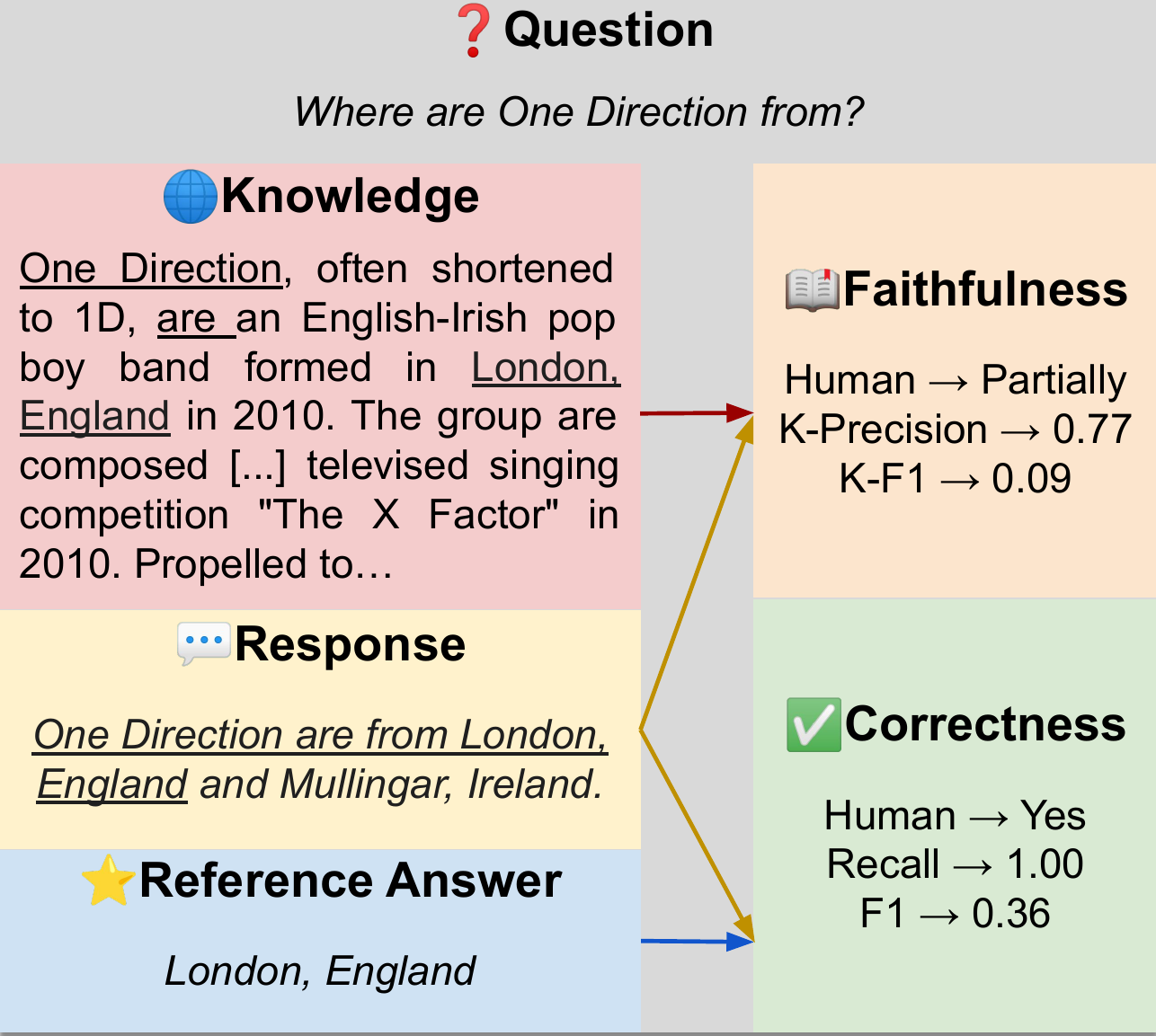}
    \caption{\small{Sample response generated by GPT-3.5. The response is correct w.r.t information need but only partially faithful w.r.t knowledge.
    Recall (\S\ref{sec:correctness_metrics}) and K-Precision (\S\ref{sec:groundedness_metrics})  approximate human judgment.}}
    \vspace{-0.5em}
    \label{fig:example_flow_diagram}
\end{figure}

Instruction-following models, such as ChatGPT, are appealing as they can perform tasks based on natural language instructions.
These models are usually trained by exposing large language models (LLMs; \citealt{brown2020gpt,zhang2022opt,touvron2023llama}) to thousands of NLP tasks formulated as instructions \cite{sanh2022t0, mishra2022naturalinstructions, wei2022flan, chung2022flant5, ouyang2022instructgpt, iyer2023optiml, touvron2023llama2} or to synthetic examples generated by other LLMs \cite{wang2022selfinstruct, Taori2023alpaca, peng2023instruction}.
In this paper, we evaluate factual question-answering (QA) ability of instruction-following models using a given set of text passages.

Instruction-following models can perform QA when provided with a task description, question, and relevant text passages
\cite{chung2022flant5}.
User-centric applications (e.g. Bing Chat) typically pair these models with a retriever or internet search to provide relevant information.
These models generate natural, informative, and verbose responses, a useful trait that helps build users' trust and engagement. However, the verbosity renders traditional evaluation metrics such as exact match (EM) and F1 unreliable, raising new challenges for evaluation \cite{kamalloo2023evaluating}.
Moreover, these models also tend to provide supplementary information that may be hallucinated \cite{chiesurin2023dangers}.

Consider \Cref{fig:example_flow_diagram}, where the user asks \textit{Where are One Direction from?}. Comparing the reference answer \textit{London, England} with the first part of model response \textit{One Direction are from London, England} yields 0 EM and 0.5 F1 score, despite both answers being effectively equivalent (the entire response scores 0.36 F1). Moreover, the model asserts that One Direction is from \textit{Mullingar}. While correct, this fact is unsupported by the provided knowledge. As EM and F1 only compare with reference answers, they cannot estimate if the model response is supported by the provided knowledge.

\revision{We posit that an optimal model should not only \textit{correctly} respond to user queries but also be \textit{faithful}, i.e. it should only disseminate information that is inferrable or directly stated by external documents \cite{rashkin-etal-2021-increasing, dziri-etal-2022-evaluating}. The resulting interpretability builds user trust and allows for dynamic knowledge updates \cite{lewis2020rag}.}
In this work, we advocate that QA models should be evaluated along two fronts -- 1) \textit{correctness w.r.t information need}, which measures model's efficacy in satisfying a user's information needs, and 2) \textit{faithfulness w.r.t provided knowledge}, which measures a model's capability to ground factual information in provided knowledge. We evaluate several recent instruction-following models -- Flan-T5 \cite{chung2022flant5}, Alpaca \cite{Taori2023alpaca}, GPT-3.5 (sibling model of \citealt{ouyang2022instructgpt}), and Llama-2 \cite{touvron2023llama2} -- on three popular factual information-seeking QA datasets -- Natural Questions (NQ; \citealt{kwiatkowski2019nq}), HotpotQA \cite{yang2018hotpotqa}, and TopiOCQA \cite{adlakha2022topiocqa}. We conduct a human analysis of 1800 model responses and correlate them with several automatic metrics for correctness and faithfulness.

Our findings suggest that for correctness, \textit{recall} -- proportion of tokens in the reference answer that are also in the model's response -- exhibits a higher correlation than traditional QA metrics like EM or F1. For faithfulness, \textit{K-Precision} -- proportion of tokens in model response that appear in the knowledge snippet -- correlates better than any other lexical metric. Although GPT-4 as an evaluator achieves the highest correlation for both correctness and faithfulness, it is expensive and prone to systematic biases \cite{wang2023llmfaireval}. We demonstrate that our proposed lexical metrics are close to GPT-4-based evaluation, allowing us to evaluate instruction-following models at a large scale.

A faithful model should not only answer when the provided knowledge is relevant, but also abstain from answering when it is irrelevant. Hence, we also consider the model's ability to abstain from answering as a measure of its faithfulness.

To summarize, our contributions are as follows:

\begin{itemize}[leftmargin=6mm]
\setlength\itemsep{0em}
    \item We annotate responses from instruction-following models for QA along the dimensions of correctness and faithfulness, and evaluate several evaluation metrics. We also analyze the nature of responses where current metrics fail.
    \item Guided by our analysis of traditional QA metrics' shortcomings, we propose simple token-overlap metrics -- \textit{recall} for correctness and \textit{K-Precision} for faithfulness -- and demonstrate their strong correlation with human judgments. 
    \item We evaluate four instruction-following models across three diverse QA tasks. Our results indicate that these models, even without any further training, are comparable to task-specific fine-tuned models for correctness. However, they struggle to be faithful to provided knowledge, often failing to accurately identify its relevance.
\end{itemize}

%% file: Sections/related_work.tex
\section{Related Work}
\paragraph{Instruction-Following Models}
These models are often trained on many NLP tasks verbalized in the form of natural language instructions \cite{wang2022super, mishra2022naturalinstructions, chung2022flant5, iyer2023optiml}.
The number of tasks varies from few tens (62 in \citealt{wei2022flan}) to several hundreds (1800+ in \citealt{iyer2023optiml}). To increase diversity and scope of NLP tasks, InstructGPT \cite{ouyang2022instructgpt} and Llama-2 \cite{touvron2023llama2} incorporate high-quality expert annotations during training. They are further trained using human feedback to align them with human preferences (RLHF; \citealt{christiano_deep_2017}).
Another popular approach, \textit{self-instruct} \cite{wang2022selfinstruct} reduces dependency on human-authored instructions by bootstrapping an LLM to generate instructions and demonstrations of new tasks. The resultant dataset is used to train instruction-following models \cite{Taori2023alpaca, peng2023instruction}.

Recent works \cite{lazaridou2022internetaugmentedlms, shi2023replug} have paired retrievers with few-shot language models for QA, alleviating the need to learn additional parameters. In contrast to these works, we evaluate retrieval-augmented instruction-following models \textit{without} any demonstrations. In these settings, models do not follow the distribution of reference answers, raising new challenges for evaluation.

\vspace{-0.5em}
\paragraph{Evaluation in  QA}
Previous research on information-seeking QA has primarily relied on lexical matching metrics such as exact match (EM) and F1 for model evaluation \cite{rajpurkar-etal-2016-squad, reddy-etal-2019-coqa}. As simple token-overlap metrics cannot capture semantic equivalence \cite{min2021efficientqa}, subsequent model-based metrics employ contextualized embeddings~\citep{zhang2020bertscore} or specialized classifier \citep{bulian2022bem} to predict equivalence. More recently, several works resort to prompting LLMs like GPT-4 \cite{openai2023gpt4} to act as evaluators \cite{chiang2023vicuna,peng2023instruction,chiang2023llmeval,kamalloo2023evaluating, liu2023geval}.

Recently, \citet{kamalloo2023evaluating} compare the correctness of InstructGPT \cite{ouyang2022instructgpt} with fine-tuned models for QA. They highlight the shortcomings of traditional QA metrics and propose model-based evaluation as a viable alternative. In contrast, we evaluate both correctness and faithfulness of instruction-following models and propose token-overlap metrics that correlate highly with human judgments.

\vspace{-0.5em}
\paragraph{Evaluating Faithfulness}
Conversational models produce factually incorrect or unsupported statements \cite{rashkin-etal-2021-increasing, dziri2022hallucination}, known as \textit{hallucinations}.
Several metrics have been proposed to detect hallucination, or conversely, to measure \textit{faithfulness}.
 Knowledge-F1 (K-F1; \citealt{shuster2021retrieval}) computes token-overlap F1 between model response and the provided knowledge. $Q^2$ \cite{honovich_q2_2021} checks for factual consistency based on automatic question generation and question answering. FaithCritic \cite{dziri_faithdial_2022} uses a trained model to predict hallucinations. 

Recently, \citet{chiesurin2023dangers} demonstrated that retrieval-augmented GPT-3 is likely to produce responses that appear trustworthy but are unfaithful to the retrieved passages. They propose K-F1++, a variant of K-F1 that discounts the tokens in the model response that appear in the question. 
In our experiments, we observe that this metric doesn't correlate well with human judgments.


\vspace{-0.5em}
\paragraph{Evaluation of Instruction-following Models} Instruction-following models have challenged previously established evaluation protocols for many NLP tasks. \citet{goyal2022zeroshotnews} demonstrate that humans prefer summaries generated by GPT-3 \cite{brown2020gpt} over fine-tuned models, however, existing automatic metrics cannot capture this preference.
\citet{xu2023lfqaevaluation} advocate multi-faceted evaluation for long-form QA that focuses on fine-grained aspects such as completeness and ease of understanding. In this work, we propose multi-faceted evaluation for factual information-seeking QA along correctness and faithfulness.


%% file: Sections/setup.tex
\section{Experimental Setup}
\label{sec:setup}

\begin{table}[t!]
\centering
\small
\begin{tabular}{lrrr}
    \toprule
    Dataset & \# Questions & Answer & \# Passages \\
    & & length & (millions) \\
    \midrule
    Natural Qns. & 3,610 & 2.16 & 21 \\
    HotpotQA & 7,405 & 2.46 & 5.2  \\
    TopiOCQA & 2,514 & 10.98 & 25.7 \\
    \bottomrule
\end{tabular}
\caption{Statistics for datasets. We use the validation splits as the test sets are hidden. Answer length is the average number of words.}
\vspace{-0.5em}
\label{tab:dataset_stats}
\end{table}

\subsection{Tasks}
\label{sec:tasks}
We evaluate instruction-following models on three diverse information-seeking QA tasks based on Wikipedia.
For each task, we use a representative popular dataset.
We describe the tasks below.
\vspace{-0.5em}
\paragraph{Open-domain QA}
Here we test the model's ability to answer questions with genuine information-seeking intent and whose answer is present in one of the Wikipedia passages. We use the open version \cite{lee2019orqa} of Natural Questions (NQ; \citealt{kwiatkowski2019nq}), that contains queries from Google search engine.
\vspace{-0.5em}
\paragraph{Multi-hop QA}
Here we test the model's ability to answer questions that require at least two Wikipedia passages to reason upon jointly.
We use HotpotQA \cite{yang2018hotpotqa} for this task.
\vspace{-0.5em}
\paragraph{Conversational QA}
Here we test the model's ability to answer questions in conversational context and whose answer is present in one of the Wikipedia passages.
We use TopiOCQA \cite{adlakha2022topiocqa}, a dataset for open-domain information-seeking dialogue.

\Cref{tab:dataset_stats} lists the total number of questions, average answer length, and the total number of passages in the Wikipedia corpus for each dataset.
These datasets contain short-form answers as they are easier and more consistent to annotate by humans.
However, as users, humans prefer verbose answers \cite{chiesurin2023dangers}.
This mismatch makes our evaluation setting realistic and important.

\subsection{Instruction-following Models}
\label{sec:instruction-tuned}
As shown in \Cref{fig:generic_prompt_template}, we use a standardized prompt template that contains an instruction, passage(s) from an information source, and the question to elicit answers from instruction-following language models. We replace the question with conversation history for TopiOCQA. Inspired from \citet{mishra2022naturalinstructions}, we formulate the instruction as -- ``\texttt{Please answer the following question given the following passages}''. We refer to this instruction as \textbf{Instr. v1}.
We consider four models that differ primarily based on their training regimes. We use the same generation parameters for all instruction-following models, described in \Cref{appendix:generation_params}.

\vspace{0.4em}
\noindent\textbf{Flan-T5} \cite{chung2022flant5} --
We use the 11B parameter obtained by training T5 \citep{raffel2020t5} on multiple instruction-following datasets \citep{sanh2022t0, wang2022super, wei2022flan}. These datasets encompass 1800+ tasks, of which 200+ are QA tasks. The training splits of NQ and HotpotQA are included in these datasets.

\vspace{0.4em}
\noindent\textbf{Alpaca} --
\citet{Taori2023alpaca} train LLaMA \citep{touvron2023llama} on GPT-3-generated demonstrations using the \textit{self-instruct} framework \cite{wang2022selfinstruct}. We use the 7B variant.

\vspace{0.4em}
\noindent\textbf{GPT-3.5}
We use the \textit{turbo} version of GPT-3.5\footnote{\href{https://openai.com/blog/introducing-chatgpt-and-whisper-apis}{openai.com/blog/introducing-chatgpt-and-whisper-apis}} described
as a sibling to the InstructGPT model \cite{ouyang2022instructgpt}. It is trained with user data from the OpenAI API and expert annotations, however, the exact distribution of training tasks and datasets is not publicly available.

\vspace{0.4em}
\noindent\textbf{Llama-2}
We use the 7B chat version of Llama-2 \cite{touvron2023llama2}. The model is initially bootstrapped on similar datasets as Flan-T5, followed by fine-tuning on dialogue-style
instructions. 

\begin{figure}[t!]
\begin{mdframed}[backgroundcolor=gray!5]
    \small
    \texttt{\{Instruction\}\\\\
    - title: \{Passage title\}\\{\{Passage text\}}\\\\
    - title: \{Passage title\}\\{\{Passage text\}}\\
    ... \\
    Question: {\{Question\}}\\
    Answer: }
\end{mdframed}
\caption{Prompt template used for evaluating instruction-following models. The passages are provided by the retriever when evaluating for correctness w.r.t information need.}
\vspace{-0.5em}
\label{fig:generic_prompt_template}
\end{figure}


\subsection{Retrieval}
\label{sec:retrieval}
To evaluate instruction-following models for correctness, we pair them with a retriever that provides the model with passages relevant to the user query. For each task, we employ a task-specific variant of DPR (Dense Passage Retrieval; \citealt{karpukhin2020dense}).
For NQ, we use a pre-trained checkpoint from \citet{karpukhin2020dense}, which is trained on multiple QA datasets. For HotpotQA, we adopt the iterative multi-hop DPR variant by \citet{xiong2021mhdr} that selects passages based on the query and prior retrievals. For TopiOCQA, we utilize the checkpoint provided by \citet{adlakha2022topiocqa} that is trained for conversational QA task.

The number of retrieved passages passed to instruction-following models is constrained by their input context size. For a fair comparison, we provide the same number of retrieved passages to each model within a specific task -- 8 for NQ and HotpotQA, and 4 for TopiOCQA.

%% file: Sections/correctness.tex
\section{Correctness w.r.t Information Need}
\label{sec:correctness}

In this section, we investigate the correctness of instruction-following models. \revision{We consider a model response to be correct if it accurately satisfies the user's information need. For example, while answering \textit{What is the capital of Canada?}, the model should convey that \textit{Ottawa} is the capital of Canada. While the model's response might include additional information like Ottawa's population, we limit the evaluation of correctness to the part of the model response that is directly relevant to the user's information need. We address evaluation of additional information in the next section (\Cref{sec:groundedness}).}
We describe lexical and semantic similarity metrics for the task in \S\ref{sec:correctness_metrics}. Next, we conduct human evaluation and compare several evaluation metrics (\S\ref{sec:human_eval_correctness}). Finally, using metrics that correlate highly with human judgments, we evaluate instruction-following models for correctness (\S\ref{sec:correctness_results}).


\subsection{Evaluation Metrics}
\label{sec:correctness_metrics}
Evaluating correctness in QA involves comparing model responses to human-annotated gold answers. Below, we describe the two broad categories of automatic evaluation metrics:

\vspace{-0.5em}
\paragraph{Lexical Matching} These metrics score a model response based on its token overlap with the gold answer. 
While some metrics perform bag-of-words matching (e.g., Exact Match (EM), F1), others consider the order of the tokens by $n$-gram matching such as METEOR \citep{banerjee2005meteor}, and ROUGE \citep{lin2004rouge}.

In this work, we also consider \textit{Recall} -- the proportion of tokens in the reference answer that are present in the model response. Recall does not penalize a verbose response, as long as it contains the reference answer tokens. Recent works \cite{Liu023lostinmiddle, Mallen2023llm_memorization} have used a similar metric whereby a model response is correct if it fully contains the reference answer as a substring. We refer to this metric as \textit{Recall (Strict)}, as it is a stricter version of token-level recall.

\vspace{-0.5em}
\paragraph{Semantic Similarity} Unlike lexical metrics that face the \textit{strictness} issues \citep{min2021efficientqa}, semantic similarity metrics typically leverage a model to predict semantic equivalence. BERTScore (\textit{BertS}, \citealt{zhang2020bertscore})
uses contextual BERT embeddings to compute precision, recall, and F1 between the model and reference answers. \textit{BEM} (BERT matching, \citealt{bulian2022bem}) employs a trained BERT model to predict the semantic equivalence based on the question, the reference answer, and the model response. We extend BEM to conversational QA task by providing the most recent question in the conversation as input.

Based on recent works \cite{chiang2023vicuna, peng2023instruction}, we also consider prompting LLMs (referred to here as \textit{GPT3.5-Eval} and 
\textit{GPT4-Eval}) to act as evaluation agents. Given the question, the reference answer, and the model response, the LLM is instructed to predict if the model response is correct or not. The prompt template and the instruction used are described on our project page\footnote{\href{https://github.com/McGill-NLP/instruct-qa/\#correctness}{github:instruct-qa:correctness}}.
\subsection{Human Evaluation}
\label{sec:human_eval_correctness}


\input{Tables/correctness_human_eval_metrics}

To establish a basis for comparing evaluation metrics, we conduct human evaluation on a subset of responses from all instruction-following models. Specifically, we focus on cases where the gold passage is included in retrieved passages. Therefore, any inaccuracies in the response can be attributed to model's failures, rather than inaccurate retrieval. For each of the three tasks, we take 100 samples, resulting in 1200 samples for the four models.

In our evaluation setup, the annotator is presented with the question (or conversation history), the reference answer, and the anonymized model response. The annotator's task is to assess if the model response is \textit{correct}, i.e., it is factually accurate and satisfies the information need underlying the user's query. We hired four NLP graduate students for the task. Each sample is labeled by two annotators, achieving an inter-annotator agreement of 92.42\% \revision{and a Fleiss' Kappa score of 76.4\%}.  In cases of disagreement, a third annotation is collected and majority vote is taken.

Out of 1200 model responses, 961 were judged correct by humans, and 239 as incorrect. \revision{In \Cref{tab:correctness_human_eval_metrics}, we report distributional statistics of scores assigned by EM, F1, Recall, and GPT4-Eval for both correct and incorrect responses.} While EM assigns a 0.0 score to almost all human-judged incorrect responses, it assigns 1.0 to only 22\% of human-judged correct responses.
F1 does only slightly better, obtaining 39\% accuracy on human-judged correct responses (when we consider responses with $\geq$ 0.5 score as correct).
This highlights the well-known strictness problem \cite{min2021efficientqa, kamalloo2023evaluating} of traditional QA metrics. In contrast, GPT4-Eval and Recall offer a relatively
balanced assessment.

\vspace{-0.5em}
\paragraph{Qualitative Analysis of Failure Cases}
\label{paragraph:correctness_error_analysis}
As evident from \Cref{tab:correctness_human_eval_metrics}, traditional QA metrics like EM and F1 tend to produce higher rate of false negatives than false positives. For instance, 78\% of answers deemed correct by humans were falsely marked incorrect (false negative) by EM, whereas only 0.4\% (1 out of 239) of answers judged incorrect by humans were marked correct (false positive). Similarly F1 has 39\% false negative rate and 3.8\% false positive rate. As the rate of false positives is extremely low and they do not disproportionately impact instruction-following models, our analysis focuses solely on the false negatives.

We analyze the models' responses that have $\leq$ 0.3 F1 score, but have been deemed correct by the annotators. This results in 448 samples out of 1200. Our classification of errors is inspired from \citet{kamalloo2023evaluating} and \citet{min2021efficientqa}, modified to focus on instruction-following models.
We list the categories of our classification and their descriptions below.

\begin{figure}[t!]
    \centering
    \includegraphics[trim={2.3cm 1.4cm 0 1.3cm},clip,width=\linewidth]{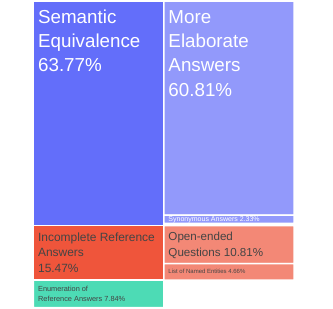}
    \caption{Categorization of failure cases when human judged the response correct and F1 $\leq$ 0.3. \textit{More Elaborate Answers} is most common failure sub-category, followed by \textit{Open-ended Questions}. }
    \vspace{-0.5em}
    \label{fig:correctness_error_analysis}
\end{figure}

\vspace{-0.5em}
\begin{itemize}[leftmargin=4mm]
    \item \textbf{Semantic Equivalence:} Here, the model response is semantically similar to the reference answer. Sub-categories include \textbf{Multinominal entities}, e.g., \textit{John Kennedy} and \textit{John F Kennedy}, \textbf{More Elaborate Answers}, e.g., \textit{yes} and \textit{yes, he is member of the band} and \textbf{Synonymous Answers}, e.g., \textit{from India} and \textit{Indian nationality}.
    \item \textbf{Symbolic Equivalence:} This primarily refers to different possible representations of numeric quantities, e.g. \textit{four seasons} and \textit{4 seasons}.
    \item \textbf{Intrinsic Ambiguity in Questions:} This refers to queries with multiple valid interpretations, leading to a range of correct answers, e.g. \textit{Who won NFL football coach of the year?} could have different answers dependent on the specific point in time being referenced.
    \item \textbf{Granularity Discrepancies:} The level of specificity in the model's response may not align with that in the reference answer. This discrepancy in granularity can be \textbf{Temporal}, e.g., \textit{August 25, 1939} and \textit{1939}, or \textbf{Spatial}, e.g., \textit{Vancouver} and \textit{British Columbia, Canada}.
    \item \textbf{Incomplete Reference Answers:} This refers to cases when the reference answers fail to cover the entire spectrum of correct responses. We consider two sub-categories -- \textbf{List of named entities} which includes questions like the cast of a movie or members of the band, and \textbf{Open-ended questions}  which included questions that can be answered in multiple different ways, all of which are not captured by reference answers., e.g., \textit{What was the Watergate scandal?}.
    \item \textbf{Enumeration of Reference Answers} is an error category where the question seeks a list (e.g., \textit{All states in north-east USA}), but each reference answer contains only one entity (e.g., \textit{``Vermont''}, \textit{``Maine''}). Instruction-following models often list multiple entities together in the response (e.g., \textit{Vermont and Maine}), leading to mismatches. This error category is very frequent in NQ.
    \item \textbf{Satisfactory Subset Response} represents the inverse, where the model's answer, though shorter than the reference, still addresses the query. An example is when a query asks for songs of an artist, the reference lists 5-6, but the model responds with only 1-2 song names (primarily seen in TopiOCQA).
\end{itemize}

\input{Figures/correctness_manual_inspection}

\noindent\Cref{fig:correctness_error_analysis} displays the distribution of error cases based on our classification. A significant portion of the errors (60.81\%) fall under the \textit{More Elaborate Answers} category. This suggests that traditional QA metrics often penalize models unjustly due to the verbose nature of their responses. The next most common sub-category, \textit{Open-ended Questions} (10.81\%), suggests that models are occasionally penalized for providing correct responses that were not included in the reference answers.

In \Cref{fig:manual_inspection}, we provide qualitative examples of \textit{More Elaborate Answers} and \textit{Open-ended Questions} sub-categories. Recall can act as an effective fix for \textit{More Elaborate Answers}. However, both lexical and semantic similarity metrics struggle with \textit{Open-ended Questions}. We also observed that this is also the most common failure sub-category for GPT4-Eval.

Overall, the results of our human evaluation and analysis indicate that traditional metrics such as EM and F1, typically used for QA models, are not well-aligned with the verbose nature of instruction-following models. To determine more suitable metrics for these models, we analyze the correlation of each metric with human assessments.



\vspace{-0.2em}
\paragraph{Correlation Between Automatic Metrics and Human Judgement}

\revision{
With only four models in our setup, correlations computed between model rankings obtained from metrics and human judgments are not statistically significant. To overcome this issue, we directly compare the human judgments with the metric scores across 1200 annotated examples.
We utilize Spearman's $\rho$ and Kendall's $\tau$, both of which have mechanisms which prevent the ties from artificially inflating or deflating the correlation measurement. For instance, we use $\tau$-$b$ for Kendall \citep{kendall-tau-b}, which adjusts the normalizing factor by discounting the number of tied pairs. Similarly, for Spearman's $\rho$, the average of the ranks that would have been assigned to all the tied values is assigned to each value.
Simply put, a metric achieves high correlation if it assigns a higher score to samples deemed correct by humans than to those deemed incorrect.}
\Cref{tab:correctness_human_correlations} presents the Spearman's $\rho$ and Kendall's $\tau$ correlation of different metrics with human judgments.
Apart from metrics detailed in \Cref{sec:correctness_metrics}, we include token-level precision, as well as precision and recall as computed using BERTScore.

\input{Tables/overall_correlations_correctness}

Notably, GPT4-Eval has the highest agreement with human judgments, with 67.47 for both Spearman and Kendall correlation, closely followed by GPT3.5-Eval.
We speculate that the language comprehension capabilities and inherent world knowledge embedded in LLMs help them overcome many of the challenges associated with evaluating responses of instruction-following models that we identified in our human evaluation study.

After GPT4-Eval and GPT3.5-Eval, Recall achieves the highest correlation with human judgment. This simple token-overlap metric correlates better than other lexical metrics or more complex semantic similarity metrics like BERTScore and BEM, likely because it does not penalize the additional verbosity in model responses.


Although LLM-based evaluations such as GPT4-Eval and GPT3.5-Eval exhibit the highest correlation with human judgments, they also have certain limitations. Accessing these proprietary models incurs substantial API costs, which renders them impractical for automatic evaluation on large-scale datasets. Moreover, the reliability of LLMs as evaluators is still unclear, as recent studies have shown that they may exhibit systematic bias \cite{wang2023llmfaireval}. Given these considerations, we rely on Recall to evaluate model performance.

\input{Tables/main_results_correctness}
\subsection{Evaluating the Correctness of Instruction-following Models}
\label{sec:correctness_results}
Equipped with proper evaluation metrics, we evaluate instruction-following models for correctness across three QA tasks. Specifically, we investigate the performance of current instruction-following models in comparison to models that have been specifically fine-tuned for those tasks.

To compare against instruction-following models, we select FiD \cite{izacard2021fid} as our task-specific fine-tuned baseline. This T5-based \cite{raffel2020t5} encoder-decoder model separately encodes each retrieved passage with the query, resulting in a set of vectors. The decoder then autoregressively generates the answer by attending to input passages and previously generated tokens. For NQ and TopiOCQA, we use the publicly available FiD checkpoints. For HotpotQA, we train our own variant using the default hyperparameters. All checkpoints are \texttt{base} variants that contain 220 million trainable parameters.

Unlike instruction-following models (\Cref{sec:retrieval}), FiD is not restricted by input context size for number of retrieved passages. We use the default settings for each dataset -- 100 passages for NQ, 50 for TopiOCQA, and up to 18 for HotpotQA. 

In \Cref{tab:main_results_correctness}, we report EM, F1, and Recall for assessing correctness. Unsurprisingly, FiD, which is fine-tuned separately on each dataset and thus emulates the distribution of reference answers, scores higher than instruction-following models on traditional QA metrics like EM and F1 (with the exception of Flan-T5 on HotpotQA). However, based on our findings (\Cref{sec:human_eval_correctness}), we rely on Recall for a more accurate evaluation. Using recall, the performance gap narrows significantly, with some instruction-following models even outperforming FiD. Notably, GPT-3.5 outperforms FiD across all three QA tasks -- by 6.48\% in NQ, 10.27\% in HotpotQA, and 9.33\% in TopiOCQA, whereas Llama-2 outperforms FiD on two out of three tasks, matching FiD's performance in TopiOCQA.
It is also worth noting that TopiOCQA serves as a test for generalization as it is not included in any instruction-tuning datasets and was released after the knowledge cutoff of GPT-3.5 (September 2021).

Overall, these results suggest that in retrieval-augmented settings, instruction-following models are equally, or sometimes even more capable than task-specific fine-tuned generators for generating correct responses w.r.t user information needs.

\revision{
We also investigate the impact of retriever on the final performance of instruction-following models, using HotpotQA as a testbed (\Cref{appendix_sec:impact_of_retriever}). Our findings underscore the importance of selecting task-specific retrievers to maximize performance.}

%% file: Tables/correctness_human_eval_metrics.tex
\begin{table}[t!]
\centering
\footnotesize
\revision{
\begin{tabular}{llcccc}
    \toprule
              &     &    EM &    F1 &  Recall &  GPT4 \\
    \midrule
    \multirow{3}{*}{Human=1 $\uparrow$} & Avg. &  0.22 &  0.45 &    0.85 &       0.89 \\
              & Median &  0.00 &  0.33 &    1.00 &       1.00 \\
              & SD &  0.42 &  0.36 &    0.30 &       0.31 \\
    \midrule
    \multirow{3}{*}{Human=0 $\downarrow$} & Avg. &  0.00 &  0.10 &    0.23 &       0.13 \\
              & Median &  0.00 &  0.00 &    0.00 &       0.00 \\
              & SD &  0.06 &  0.16 &    0.32 &       0.33 \\
    \bottomrule
    \end{tabular}
}
\caption{\revision{Average, Median and standard deviation (SD) of scores by evaluation metrics when humans judge the model response correct (1) vs incorrect (0). EM and F1 tend to be strict whereas Recall and GPT4-Eval are more balanced.}}
\vspace{-0.5em}
\label{tab:correctness_human_eval_metrics}
\end{table}

%% file: Figures/correctness_manual_inspection.tex
\begin{figure}[t!]
\small
\begin{tabular}{p{\columnwidth}}
\toprule
\textbf{Error Type:} More Elaborate Answers (60.81\%)\\
\midrule
\textbf{Dataset}: HotpotQA \\
\textbf{Question}: At what theater is the composer and lyricist for the musical Big Fish a residential artist? \\
\textbf{Ref. Answer}: Ars Nova Theater \\
\textbf{Response (Alpaca)}: The composer and lyricist for the musical Big Fish, Andrew Lippa, is a residential artist at the Ars Nova Theater in New York City.\\
\toprule
\textbf{Error Type:} Open-ended Questions (10.81\%)\\
\midrule
\textbf{Dataset}: TopiOCQA \\
\textbf{Question}: what was the watergate scandal? \\
\textbf{Ref. Answer}: It was an array of clandestine and often illegal activities undertaken by members of the Nixon administration. \\
\textbf{Response (Flan-T5)}: The Watergate scandal was a political scandal in the United States involving the administration of U.S. President Richard Nixon from 1972 to 1974 that led to Nixon's resignation.\\

\bottomrule
\end{tabular}
\caption{Qualitative examples of two most common failure modes of F1 -- \textit{Open-ended Questions} and \textit{More Elaborate Answers}.}
\label{fig:manual_inspection}
\end{figure}

%% file: Tables/overall_correlations_correctness.tex
\begin{table}[t!]
\centering
\footnotesize
\begin{tabular}{lcc}
\toprule
           Metric &      Spearman $\rho$&       Kendall $\tau$\\
\midrule
               EM &        27.326 &        27.326 \\
               F1 &        47.341 &        40.164 \\
           Recall & \textbf{60.048} & \textbf{55.622} \\
       Recall (S) &        52.535 &        52.535 \\
        Precision &        43.929 &        37.636 \\
           METEOR &        48.232 &        39.845 \\
          Rouge-L &        45.874 &        38.797 \\
       \midrule
BertS (F1) &        31.853 &        26.133 \\
   BertS (Recall) &        38.841 &        31.866 \\
BertS (Precision) &        20.876 &        17.129 \\
              BEM &        53.704 &        43.868 \\
      GPT3.5-Eval &        61.353 &        61.353 \\
        GPT4-Eval & \textbf{67.474} & \textbf{67.474} \\
\bottomrule
\end{tabular}

\caption{Correlation of several automatic evaluation metrics with human judgments for correctness w.r.t information need. GPT4-Eval achieves the highest correlation overall. Recall is the highest correlated among all lexical metrics.}
\vspace{-0.5em}
\label{tab:correctness_human_correlations}
\end{table}

%% file: Tables/main_results_correctness.tex
\begin{table}[t!]
\centering
\footnotesize
\begin{tabular}{ll|ccc}
\toprule
 & Model         &               EM $\uparrow$ &               F1 $\uparrow$ &           Recall $\uparrow$ \\
\midrule
\parbox[t]{2mm}{\multirow{5}{*}{\rotatebox[origin=c]{90}{NQ}}} & FiD &     \textbf{46.57} &     \textbf{53.93} &            54.45 \\
         & Flan-T5 &  \underline{41.16} &  \underline{50.62} &            54.03 \\
         & Alpaca &             8.84 &             19.5 &            48.82 \\
         & GPT-3.5 &             1.41 &            16.22 &  \underline{57.98} \\
         & Llama-2 &             0.64 &             9.78 &     \textbf{59.28} \\
\midrule
\parbox[t]{2mm}{\multirow{5}{*}{\rotatebox[origin=c]{90}{HotpotQA}}} & FiD &  \underline{48.43} &  \underline{60.16} &            60.55 \\
         & Flan-T5 &     \textbf{58.12} &     \textbf{71.14} &     \textbf{71.28} \\
         & Alpaca &            16.27 &            33.45 &            57.53 \\
         & GPT-3.5 &             5.66 &            22.49 &            66.77 \\
         & Llama-2 &             1.39 &            15.15 &  \underline{69.75} \\
\midrule
\parbox[t]{2mm}{\multirow{5}{*}{\rotatebox[origin=c]{90}{TopiOCQA}}} & FiD &     \textbf{36.48} &     \textbf{58.52} &  \underline{61.64} \\
         & Flan-T5 &  \underline{18.34} &  \underline{43.17} &            52.54 \\
         & Alpaca &             5.85 &            26.72 &            43.37 \\
         & GPT-3.5 &              2.7 &            34.32 &     \textbf{67.39} \\
         & Llama-2 &             0.95 &            22.79 &             61.4 \\
\bottomrule
\end{tabular}

\caption{Comparison of instruction-following models with FiD for correctness. EM and F1 rank FiD higher on NQ and TopiOCQA. According to Recall, which is more correlated with human judgments, GPT-3.5 outperforms FiD on all three datasets.}
\vspace{-0.5em}

\label{tab:main_results_correctness}
\end{table}

%% file: Sections/groundedness.tex
\section{Faithfulness w.r.t Provided Knowledge}
\label{sec:groundedness}

Instruction-following models often provide verbose responses with additional information apart from user information needs.
For example, when asked \textit{What is the capital of Canada?}, the model might add information about the population -- \textit{Ottawa is the capital of Canada, with a population of 1,017,449}.
\revision{Evaluating the correctness of this supplementary information is challenging without an oracle. Therefore, we focus on a more limited goal of \textit{faithfulness} \cite{Rashkin2021attribution, dziri2022hallucination, chiesurin2023dangers}, which measures if the supplementary information is inferable or directly stated in the knowledge provided as input to these models. A faithful model helps build user trust and enables knowledge configurability.}


Following \citet{dziri_faithdial_2022}, we posit that a faithful model response should be fully grounded in the provided knowledge. Based on this hypothesis, we split our analysis into two parts -- 1) faithfulness w.r.t relevant knowledge, where we provide the model with the relevant gold passage and evaluate the groundedness of its response, and 2) faithfulness w.r.t irrelevant knowledge, where we provide a related but irrelevant passage and measure how often the model refuses to answer.

In this section, we first describe the automatic evaluation metrics for evaluating faithfulness (\S\ref{sec:groundedness_metrics}). Next, similar to correctness, we conduct human evaluation to identify optimal metrics for faithfulness w.r.t relevant knowledge (\S\ref{sec:groundedness_gold_passage}). Finally, after outlining our approach to evaluate a model faithfulness w.r.t irrelevant knowledge (\S\ref{sec:unanswerability}), we present the results from large-scale evaluation of instruction-following models (\S\ref{sec:faithfulness_results}).

\subsection{Evaluation Metrics}
\label{sec:groundedness_metrics}
Given the user question or the conversation history (denoted by $\mathcal{H}$), the gold passage $\mathcal{K}$, and the model response $u$, the objective of the metric is to check if $u$ can be inferred from $\mathcal{K}$.
We explore several reference-free faithfulness and groundedness metrics in the literature, broadly categorized into two:
\vspace{-0.6em}
\paragraph{Lexical Matching}
Knowledge-F1 (denoted \textbf{K-F1}) is a lexical overlap metric widely used for knowledge-grounded dialogue \cite{shuster2021retrieval, dziri_faithdial_2022} that checks for F1 overlap between the tokens of $u$ and $\mathcal{K}$. As K-F1 checks for \textit{equivalence} between the model response and the knowledge snippet, we argue that it is unsuitable for information-seeking QA tasks. Grounding $u$ in $\mathcal{K}$ in these tasks is an inherently asymmetric task, i.e., $u$ can be a subset of $\mathcal{K}$ but $\mathcal{K}$ cannot be a subset of $u$. To capture this intuition, we propose \textbf{K-Precision} -- the proportion of tokens in the model response $u$ that are present in $\mathcal{K}$.

\citet{chiesurin2023dangers} propose K-F1++, a variant of K-F1 that discounts tokens from user question or the conversation history in the model response.
We also consider K-Precision++, which applies similar discounting.

\vspace{-0.5em}
\paragraph{Semantic Similarity} A parallel to K-F1 in semantic space, we explore using BERTScore to measure semantic similarity between $\mathcal{K}$ and $u$ based on contextual BERT token embeddings (denoted \textbf{K-BertS}). We also consider \textbf{FaithCritic}, a hallucination critic model by \citet{dziri_faith_2023} that evaluates whether a response entails a given passage.
\textbf{$\mathbf{Q^2}$} \citep{honovich_q2_2021} is another evaluation metric used to quantify factual consistency between responses and provided passages using automatic question generation, question answering, and natural language inference (NLI) models.

Similar to correctness, we investigate prompting LLMs to act as evaluators (\textbf{LLMCritic}).
More specifically, we prompt GPT-3.5 and GPT-4 to annotate whether a given response uses \emph{only} the knowledge present in the provided passage. The prompt template and the instruction used are described on our project page\footnote{\href{https://github.com/McGill-NLP/instruct-qa/\#faithfulness}{github:instruct-qa:faithfulness}}.

\subsection{Faithfulness w.r.t Relevant Knowledge}
\label{sec:groundedness_gold_passage}

We use the same template and prompt that we did for correctness (\Cref{sec:instruction-tuned}), but replace the retrieved passages with the gold passage(s).

\vspace{-0.5em}
\paragraph{Human Evaluation Setup}
We randomly sampled $50$ examples for each dataset, resulting in 600 examples across four models. For each sample, we provide annotators with the question (or the conversation history), model response, and the gold passage(s). They are given two tasks -- 1) to verify if the given passage is indeed relevant to the user's query, and 2) to determine if the model response is ``fully'', ``partially'', or ``not at all'' supported by the passages.
We retain the same annotators from the previous task. Each sample is annotated twice, and in case of disagreement, a third annotation is collected for a majority vote. The annotators achieved an inter-annotator agreement of 86.33\% \revision{and Fleiss' Kappa score of 70.57\%}. For our analysis, we filter out samples for which the passage is marked as not relevant to the query, resulting in 544 samples.

\input{Figures/faithfulness_manual_inspection}



\vspace{-0.5em}
\paragraph{Human Evaluation Results} Overall, 85.3\% responses were marked as ``fully'' supported by the gold passage, 9\% as ``partially'' and 5.7\% as ``not at all''.
We include examples of model responses marked as ``partially'' and ``not at all'' in \Cref{fig:manual_inspection_faithfulness}. Responses in the ``not at all'' category generally contain short hallucinated information, while "partially" supported responses tend to be lexically aligned with the passage but with slight modifications in some pieces of information.





\vspace{-0.5em}
\paragraph{Correlation of Automatic Evaluation Metrics}
To compare with automatic evaluation metrics, we consider model responses marked as ``fully'' as \textit{faithful}, assigning them a score of 1.0. The other two categories are given a score of 0.0.
We calculate Spearman's $\rho$ and Kendall's $\tau$ correlation between assessments of automatic metrics and human judgments and report the results in \Cref{tab:overall_correlations_faithfulness}.

We find that GPT-4-based LLMCritic correlates the most with human evaluation. K-Precision, the token-overlap metric that is invariant to the length of the knowledge snippet, is a close second, better than other semantic similarity metrics like K-BertS, FaithCritic, and $Q^2$. This indicates that models trained to detect hallucinations in knowledge-grounded dialogues do not generalize well to information-seeking QA tasks. Surprisingly, K-Precision also outperforms GPT-3.5-based LLMCritic, indicating that verifying faithfulness is still a challenging task for LLMs.

\input{Tables/overall_correlations_faithfulness}

Although GPT-4-based LLMCritic achieves the highest Spearman correlation of 54.99, it is still only moderately correlated with human judgments, indicating that accurately quantifying faithfulness is a challenging task for current evaluation metrics.
K-Precision is a simple interpretable token-overlap metric that can serve as a strong baseline for the development of more robust automatic metrics in the future. In this work, we rely on K-Precision to evaluate faithfulness w.r.t relevant knowledge.

\subsection{Faithfulness w.r.t Irrelevant Knowledge}
\label{sec:unanswerability}

An ideal model for QA should comprehend the provided passage  and avoid answering if the passage lacks relevant information.
To test this, we provide the models with a passage that is likely to be irrelevant but related (e.g., a query about Tom Cruise movie will be provided a Korean movie passage that has nothing to do with Tom Cruise).
To do so, we treat the first passage after the thousandth ranked passage as irrelevant but related.

Our preliminary experiments demonstrated that without explicit instruction, Flan-T5 and Alpaca did not refrain from answering at all. Hence, we modify the instruction from \Cref{sec:instruction-tuned} (Instr. v1) to direct the model to refrain from answering if the passage is deemed irrelevant -- \texttt{Please answer the following question given the following passages. If the answer is not in the passages or cannot be inferred from the passages, respond as ``I don't know''}. We refer to this instruction as \textbf{Instr. v2}. We report the proportion of model responses that contain \textit{I don't know} and other observed synonymous expressions, referred to as $P_{IR}$. To test for any bias this instruction modification might introduce, we also check for the model's answer abstinence when provided with the gold passage, denoted by $P_{G}$. Ideally, a model should always refrain from answering when given irrelevant information and never refrain when given the correct passage(s).

\input{Tables/main_results_faithfulness}

\subsection{Evaluating Faithfulness of Instruction-following Models}
\label{sec:faithfulness_results}

We conduct large-scale evaluation of instruction-following models for faithfulness. \Cref{tab:main_results_faithfulness} reports the results for faithfulness w.r.t both relevant and irrelevant knowledge.



\vspace{-0.5em}
\paragraph{Faithfulness w.r.t relevant knowledge} We report K-Precision, the metric most correlated with human judgments (\Cref{sec:groundedness_gold_passage}). Flan-T5 achieves the highest score for all three tasks, outperforming all other models by a significant margin. GPT-3.5 is the least faithful for NQ, while Llama-2 is the least faithful for HotpotQA and TopiOCQA.
The stark difference between the scores of Flan-T5 and other models denotes a trade-off between correctness and faithfulness -- GPT-3.5 and Llama-2 outperform Flan-T5 in correctness but lag behind significantly in faithfulness.



\vspace{-0.5em}
\paragraph{Answer refraining and prompt sensitivity}
When explicitly instructed to output ``I don’t know'' and given an irrelevant passage, GPT-3.5 most often refrains from answering (98\% in NQ and HotpotQA, 88\% in TopiOCQA), followed closely by Llama-2.
Alpaca almost always answers, indicating that it either fails to detect when the answer is absent or it has difficulty following the instruction.
While Flan-T5 successfully abstains from answering on NQ and HotpotQA, it fails on TopiOCQA, indicating that it struggles with out-of-distribution (TopiOCQA is not included in its training).

When provided with gold passage with the same instruction, surprisingly, both GPT-3.5 and Llama-2 still refrain from answering, with Llama-2 refraining to answer more than 50\% of the time across all three datasets.
This indicates further research is required for models to identify when and when not to refrain from answering.


\revision{We extend the evaluation of faithfulness to real-world scenarios, providing models with retrieved passages instead of gold or irrelevant passages. Additionally, we explore the impact of modifying the instruction (Instr. v1 vs Instr. v2), on both correctness and faithfulness (\Cref{appendix_sec:real_world_holistic_eval}).}



%% file: Figures/faithfulness_manual_inspection.tex
\begin{figure}[t!]
\small
\begin{tabular}{p{\columnwidth}}

\toprule
\textbf{1. Human Annotation:} Not at all \\
\midrule
\textbf{Question (NQ)}: When did they replace lead with graphite in pencils? \\
\textbf{Knowledge}: Pencil - 
[...] misconception that the graphite in the pencil is lead,
[...] \textcolor{teal}{even though it never contained the element lead.} [...]
\\
\textbf{Ref. Answer}: never contained the element lead \\
\textbf{Response (Flan-T5)}: \textcolor{purple}{1835}\\
\checkspace{
\textbf{Scores}: K-F1: 0.0, K-Precision: 0.0, K-BertS (F1): 76.22, $Q^2$ (F1): 0.0, $Q^2$ (NLI): 0.0, FaithCritic: 99.96, LLMCritic(GPT-3.5): 100.0, LLMCritic(GPT-3.5): 0.0\\
}

\toprule
\textbf{2. Human Annotation:} Partially \\
\midrule

\textbf{Question (TopiOCQA)}: What are dragonfly flight speeds? \\
\textbf{Knowledge}: Dragonfly (Flight speed) - 
[...] a \textcolor{teal}{maximum speed of 36–54 km/h (22–34 mph)} with {average cruising speed of about 16 km/h (9.9 mph)}. Dragonflies can \textcolor{teal}{travel at 100 body-lengths per second in forward flight}, and three lengths per second backwards
\\
\textbf{Ref. Answer}: It can travel at 100 body-lengths per second in forward flight, and three lengths per second backwards. \\
\textbf{Response (GPT-3.5)}: Dragonflies [...] \textcolor{teal}{maximum speed of around 30 miles per hour} with an \textcolor{purple}{average cruising speed of about 10-15 miles per hour}. They can also \textcolor{teal}{travel at 100 body-lengths per second in forward flight}
\\
\checkspace{
\textbf{Scores}:
K-F1: 54.54, K-Precision: 72.97, K-BertS (F1): 89.48, $Q^2$ (F1): 40.0, $Q^2$ (NLI): 40.0, FaithCritic: 99.55, LLMCritic(GPT-3.5): 100.0, LLMCritic(GPT-3.5): 100.0\\
}

\bottomrule

\end{tabular}
\caption{Examples of human annotations for faithfulness w.r.t relevant knowledge. In ``Partially'', part of the response is unsupported by the passage, while ``Not at all'' refers to completely hallucinated response. Text in \textcolor{purple}{purple} indicates hallucination;  \textcolor{teal}{teal} indicates grounded in provided knowledge.
}
\vspace{-0.5em}
\label{fig:manual_inspection_faithfulness}
\end{figure}

%% file: Tables/overall_correlations_faithfulness.tex
\begin{table}[t!]
\centering
\footnotesize
\begin{tabular}{lcc}
\toprule
             Metric &         Spearman $\rho$&          Kendall $\tau$\\
\midrule
               K-F1 &          -10.266 &           -8.397 \\
             K-F1++ &           -5.541 &           -4.537 \\
        K-Precision & \underline{49.849} & \underline{43.384} \\
      K-Precision++ &           43.559 &           39.041 \\
           K-Recall &          -13.931 &          -11.397 \\
       \midrule
K-BertS (F1) &           -0.456 &           -0.373 \\
K-BertS (Precision) &           23.083 &           18.866 \\
   K-BertS (Recall) &          -16.817 &          -13.745 \\
        FaithCritic &           11.277 &            9.218 \\
     $Q^2$ (F1) &           28.708 &           24.478 \\
    $Q^2$ (NLI) &           28.862 &           25.084 \\
LLMCritic (GPT-3.5) &           23.851 &           23.851 \\
  LLMCritic (GPT-4) &    \textbf{54.995} &    \textbf{54.995} \\
\bottomrule
\end{tabular}

\caption{Correlation of evaluation metrics with human judgements for faithfulness w.r.t relevant knowledge. LLMCritic (GPT-4) achieves the highest correlation. K-Precision is a close second.}
\vspace{-0.5em}
\label{tab:overall_correlations_faithfulness}
\end{table}

%% file: Tables/main_results_faithfulness.tex
\begin{table}[t!]
\centering
\footnotesize
\begin{tabular}{ll|c|cc}
\toprule
 & Model         &     K-Precision $\uparrow$ &  $P_{IR}\uparrow$ &       $P_{G}\downarrow$ \\
\midrule
\parbox[t]{2mm}{\multirow{4}{*}{\rotatebox[origin=c]{90}{NQ}}} & Flan-T5 &     \textbf{94.0} &            92.0 &  \underline{24.8} \\
         & Alpaca &  \underline{69.4} &             0.0 &      \textbf{0.0} \\
         & GPT-3.5 &            65.5 &     \textbf{98.4} &            47.4 \\
         & Llama-2 &            69.4 &  \underline{97.8} &            57.8 \\
\midrule
\parbox[t]{2mm}{\multirow{4}{*}{\rotatebox[origin=c]{90}{HotpotQA}}} & Flan-T5 &     \textbf{92.1} &            77.1 &   \underline{1.6} \\
         & Alpaca &  \underline{87.1} &             0.1 &      \textbf{0.1} \\
         & GPT-3.5 &            81.4 &     \textbf{98.2} &            25.5 \\
         & Llama-2 &            75.8 &  \underline{97.7} &            61.5 \\
\midrule
\parbox[t]{2mm}{\multirow{4}{*}{\rotatebox[origin=c]{90}{TopiOCQA}}} & Flan-T5 &     \textbf{86.4} &            40.8 &   \underline{7.7} \\
         & Alpaca &            66.8 &             1.3 &      \textbf{0.8} \\
         & GPT-3.5 &  \underline{69.6} &     \textbf{88.2} &            31.8 \\
         & Llama-2 &            65.4 &  \underline{79.1} &            52.4 \\
\bottomrule
\end{tabular}

\caption{Performance of instruction-following models for faithfulness. K-Precision evaluates faithfulness w.r.t relevant knowledge (\Cref{sec:groundedness_gold_passage}). $P_{IR}$ and $P_G$ denote the proportion of responses where model refrained from answering when provided with incorrect or gold passage respectively, along with a modified instruction (\Cref{sec:unanswerability}).}
\vspace{-0.5em}

\label{tab:main_results_faithfulness}
\end{table}

%% file: Sections/conclusion.tex
\section{Conclusion}
\label{sec:conclusion}

In this paper, we analyze the responses of instruction-following models for QA along the dimensions of correctness w.r.t information need and faithfulness w.r.t. provided knowledge.
Our results show that Recall and K-Precision metrics correlate well with human judgments for correctness and faithfulness respectively. 

On evaluating instruction-following models using these proposed metrics, we find that these models demonstrate a tradeoff between correctness and faithfulness. 
GPT-3.5 and Llama-2 achieve high scores for correctness but have difficulty being faithful to the provided knowledge.
Moreover, they struggle to decide when to refrain from answering.

When using instruction-following models for QA, we urge to the community to move away from reporting a single overall score and adopt a more holistic evaluation that reports correctness, faithfulness, and the ability to refrain from answering.

\vspace{-0.5em}
\paragraph{Limitations}
\revision{
Although we evaluate for correctness, faithfulness, and the ability to refrain from answering, it is not an exhaustive list of all the desirable properties of a QA model. Previous works \cite{xu2023lfqaevaluation} have focused on aspects like completeness and ease of understanding for long-form QA. We leave the evaluation of these properties for information-seeking QA to future work.

It is important to note that low faithfulness of a response does not imply that it is incorrect. The model can potentially provide accurate information using its parametric knowledge. However, such knowledge is difficult to interpret and modify.

We propose Recall and K-Precision for correctness and faithfulness respectively. Although these metrics correlate highly with human judgments, they are easy to hack.
For instance, Recall might score an affirmative statement and its negated version equally, despite their contrasting meanings. However, QA models tend to answer in affirmation rather than negation.
Similarly, K-Precision can be hacked by copying all the knowledge from the prompt. However, such strategy will be penalized heavily when evaluated for faithfulness w.r.t. irrelevant knowledge.
}

%% file: Sections/acknowledgements.tex
\section*{Acknowledgements}
We thank the reviewers and action editors for their valuable feedback.
Furthermore, we thank the members of SR’s research group for providing feedback throughout the project.
We thank Eva Portelance and Ehsan Kamalloo for helpful discussions.
PB is supported by the Mila-Intel Grant program.
XHL and NM are supported by the NSERC CGSD Fellowship.
SR is supported by a Facebook CIFAR AI Chair and NSERC Discovery Grant program.

%% file: Sections/appendix_experimental_details.tex
\section{Generation parameters for instruction-following models}
\label{appendix:generation_params}

We use the following generation parameters for all instruction-following models:
\begin{itemize}
    \setlength\itemsep{-0.3em}
    \item Top-p: $p=0.95$
    \item Temperature: $t=0.95$
    \item Seed: $s=0$
    \item Min. new tokens: $\min_{token}=1$
\end{itemize}
For GPT-3.5, we did not specify any limit for maximum new tokens. For other models, we specified $\max_{token}$ as 500 to prevent out-of-bound memory errors on our GPU infrastructure. 

%% file: Sections/appendix_impact_of_retriever.tex
\section{Impact of Retriever on Correctness}
\label{appendix_sec:impact_of_retriever}

\vspace{-0.5em}

\begin{figure}[!ht]
    \centering
    \includegraphics[width=\linewidth]{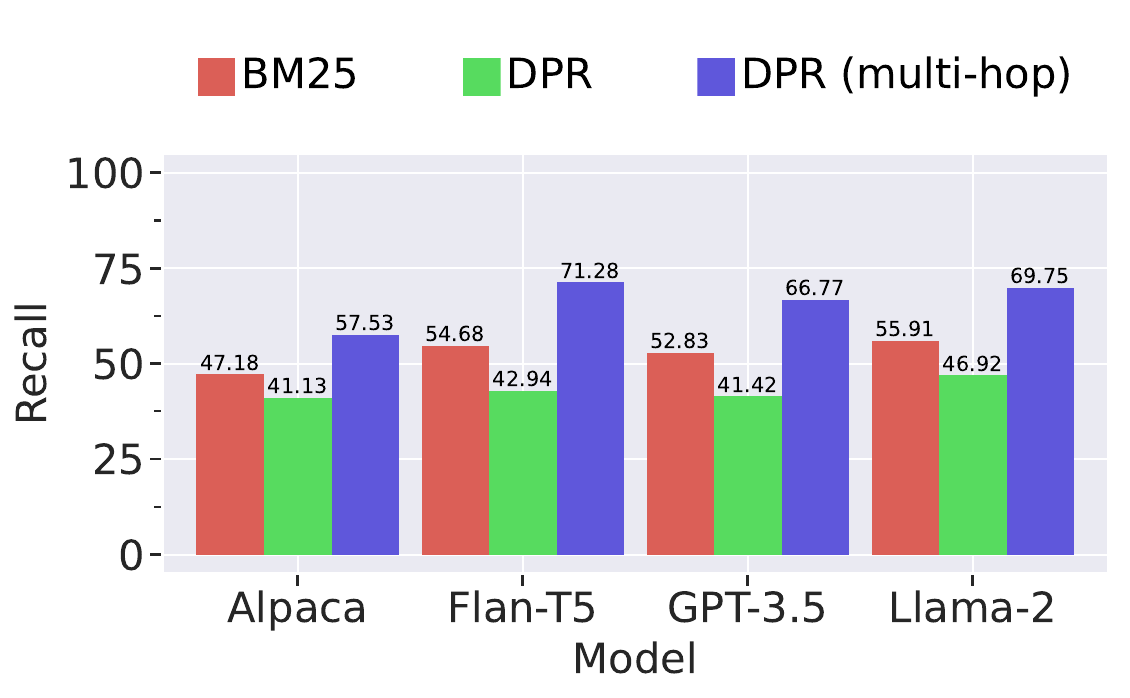}
    \caption{Correctness evaluation of instruction-following models across various retrievers for HotpotQA dataset. DPR (multi-hop), the task-specific retriever for multi-hop QA, performs the best.}
    \vspace{-0.5em}
    \label{fig:hpqa_retrievers}
\end{figure}

\input{Tables/real_world_holistic_eval.tex}

We evaluate the impact of different retrievers on the correctness of instruction-following models, using HotpotQA as a testing benchmark. We consider three retrievers -- (1) BM25 \citep{robertson1995okapi}, a sparse lexical-overlap based retriever, (2) DPR \citep{karpukhin2020dense}, a dense retriever trained on multiple QA datasets, and (3) \citet{xiong2021mhdr}, a multi-hop version of DPR trained on HotpotQA, which we refer to as \textit{DPR (multi-hop)}. We use the same prompt template and evaluation setup as \Cref{sec:correctness}. 

The comparison of different retrievers is presented in \Cref{fig:hpqa_retrievers}. For all instruction-following models, DPR (multi-hop) performs the best, highlighting the importance of using task-specific retrievers to maximize performance. Similar to \citet{sidiropoulos-etal-2021-combining}, we found that BM25 outperforms DPR across all instruction-following models, potentially due its ability to exploit high overlap between the query and gold passages.


%% file: Tables/real_world_holistic_eval.tex
\begin{table*}[ht!]
    \centering
    \footnotesize
    \begin{tabular}{ll|cc|cc|ccc}
        \toprule
               &    \multirow{2}{*}{Model}     & \multicolumn{2}{c|}{Correctness w.r.t} & \multicolumn{2}{c|}{Faithfulness w.r.t Retrieved} & \multicolumn{3}{c}{ \multirow{2}{*}{Answer Abstinence ($P_G \downarrow$)}} \\
          &    & \multicolumn{2}{c|}{Information Need (Recall $\uparrow$)} & \multicolumn{2}{c|}{Knowledge (K-Precision $\uparrow$)} &  & \\
        \midrule
                 &         &     Instr. v1 &      Instr. v2 &          Instr. v1 &      Instr. v2 &      Instr. v1 &      Instr. v2  &  \% Gold\\
        \midrule
\parbox[t]{2mm}{\multirow{4}{*}{\rotatebox[origin=c]{90}{NQ}}} & Flan-T5 &  54.03 &  \textbf{50.52} &       \textbf{96.59} &  \textbf{98.88} &  \textbf{0.00} &   5.95 & \multirow{4}{*}{36.79} \\
        & Alpaca &  48.82 &  48.00 &       80.87 &  79.16 &  \textbf{0.00} &   \textbf{0.08} & \\
        & GPT-3.5 &  57.98 &  34.10 &       81.50 &  96.27 &  2.18 &  39.01 & \\
        & Llama-2 &  \textbf{59.28} &  44.03 &       83.26 &  76.46 &  0.08 &  31.55 & \\
\midrule
\parbox[t]{2mm}{\multirow{4}{*}{\rotatebox[origin=c]{90}{HotpotQA}}}& Flan-T5 &  \textbf{71.28} &  \textbf{69.68} &       \textbf{91.18} &  92.81 &  \textbf{0.00} &   0.87 & \multirow{4}{*}{79.28} \\
        & Alpaca &  57.53 &  57.37 &       86.98 &  86.85 &  \textbf{0.00} &   \textbf{0.00} & \\
        & GPT-3.5 &  66.77 &  40.08 &       80.64 &  \textbf{93.72} &  2.42 &  41.17 & \\
        & Llama-2 &  69.75 &  47.56 &       81.75 &  67.35 &  0.22 &  61.51 & \\
\midrule
\parbox[t]{2mm}{\multirow{4}{*}{\rotatebox[origin=c]{90}{TopiOCQA}}} & Flan-T5 &  52.54 &  \textbf{49.16} &       \textbf{91.77} &  \textbf{94.41} &  0.38 &   6.52 & \multirow{4}{*}{62.25} \\
        & Alpaca &  43.37 &  42.85 &       72.57 &  67.33 &  0.19 &   \textbf{0.26} & \\
        & GPT-3.5 &  \textbf{67.39} &  46.76 &       80.80 &  90.74 &  0.64 &  25.11 & \\
        & Llama-2 &  61.40 &  42.62 &       79.57 &  70.66 &  \textbf{0.13} &  33.29 & \\
\bottomrule
\end{tabular}
    
\caption{Evaluation of correctness (Recall), faithfulness (K-Precision), and answer abstinence ($P_G$) of retrieval-augmented instruction-following models along different instructions. $P_G$ is reported on a subset of questions for which the gold passage is present in the retrieved passages, denoted by \% Gold.}
\vspace{-0.5em}
\label{tab:holistic_eval}
\end{table*}

%% file: Sections/appendix_real_world_holistic_eval.tex
\section{Evaluation of Instruction-Following Models in Real-world Settings}
\label{appendix_sec:real_world_holistic_eval}

In this section, we evaluate faithfulness of instruction-following models in real-world settings. We also investigate the impact of changing instructions on both correctness and faithfulness.

\noindent\textbf{Experiment Setup}
We follow the prompt template from \Cref{fig:generic_prompt_template}, providing models with an instruction, retrieved passages, and question or conversation history.
We consider two instructions -- Instr. v1 (\Cref{sec:instruction-tuned}) and Instr. v2 (\Cref{sec:unanswerability}). The latter adds a directive to not answer if passages are irrelevant.
For evaluation, we use Recall for correctness and K-precision for faithfulness.
The correctness results with Instr. v1 are copied from \Cref{tab:main_results_correctness} for easy comparison with Instr. v2.
With retrieved passages, K-Precision is computed as the proportion of tokens in the model's response that are present in the retrieved passages. For Instr. v2, we consider ``I don't know.'' to be an additional retrieved passage, hence marking the response as faithful if the model refrained from answering.

It is non-trivial to determine when the model should refrain from answering (\Cref{sec:unanswerability}) with retrieved passages as they may be relevant even if they are not gold passages.
However, if the gold passages are retrieved, the model should definitely \textit{not} abstain from answering. Therefore, we report $P_G$, i.e. the proportion of times model refrained from answering when the gold passage was present, as a proxy of answer abstinence (lower is better). A model that always output ``I don't know.'' with Instr. v2 will achieve the best score in K-Precision but the worst score in $P_G$.

\noindent\textbf{Results}
We report the results in \Cref{tab:holistic_eval}. Flan-T5 outperforms all other models by a significant margin for faithfulness under Instr. v1, consistent with previous findings (\Cref{tab:main_results_faithfulness}).
Switching to Instr. v2 increases the faithfulness of Flan-T5 and GPT-3.5, with GPT-3.5 showing the largest gain.
Llama-2's faithfulness drops by 12.33\% across all three tasks. Our manual inspection reveals that Llama-2 often explains its reasoning for not answering, which has a low overlap with the retrieved passages.

The instruction switch impacts the correctness of all models, except Alpaca. GPT-3.5 experiences the largest drop in performance across all three datasets (37.26\%), followed by Llama-2 (29.37\%). This decline is likely due to the models' increased tendency to refrain from answering under Instr. v2. $P_G$ scores support this hypothesis -- GPT-3.5 refrains from answering 39.01\% of the time on NQ with Instr. v2, compared to 2.18\% with Instr. v1. Overall, these scores indicate a similar observation as noted in \Cref{sec:faithfulness_results} -- GPT-3.5 and Llama-2 refrain from answering more often than needed.

Alpaca is relatively unaffected with the change in instruction across correctness, faithfulness and answer abstinence, further confirming the observation that it has difficulty following the instruction to refrain from answering.

%% file: main-5785-Adlakha.bbl
\begin{thebibliography}{55}
\expandafter\ifx\csname natexlab\endcsname\relax\def\natexlab#1{#1}\fi

\bibitem[{Adlakha et~al.(2022)Adlakha, Dhuliawala, Suleman, de~Vries, and
  Reddy}]{adlakha2022topiocqa}
Vaibhav Adlakha, Shehzaad Dhuliawala, Kaheer Suleman, Harm de~Vries, and Siva
  Reddy. 2022.
\newblock Topi{OCQA}: Open-domain conversational question answering with topic
  switching.
\newblock \emph{Transactions of the Association for Computational Linguistics},
  10:468--483.

\bibitem[{Banerjee and Lavie(2005)}]{banerjee2005meteor}
Satanjeev Banerjee and Alon Lavie. 2005.
\newblock \href {https://aclanthology.org/W05-0909} {{METEOR}: An automatic
  metric for {MT} evaluation with improved correlation with human judgments}.
\newblock In \emph{Proceedings of the {ACL} Workshop on Intrinsic and Extrinsic
  Evaluation Measures for Machine Translation and/or Summarization}, pages
  65--72, Ann Arbor, Michigan. Association for Computational Linguistics.

\bibitem[{Brown et~al.(2020)Brown, Mann, Ryder, Subbiah, Kaplan, Dhariwal,
  Neelakantan, Shyam, Sastry, Askell, Agarwal, Herbert-Voss, Krueger, Henighan,
  Child, Ramesh, Ziegler, Wu, Winter, Hesse, Chen, Sigler, Litwin, Gray, Chess,
  Clark, Berner, McCandlish, Radford, Sutskever, and Amodei}]{brown2020gpt}
Tom Brown, Benjamin Mann, Nick Ryder, Melanie Subbiah, Jared~D Kaplan, Prafulla
  Dhariwal, Arvind Neelakantan, Pranav Shyam, Girish Sastry, Amanda Askell,
  Sandhini Agarwal, Ariel Herbert-Voss, Gretchen Krueger, Tom Henighan, Rewon
  Child, Aditya Ramesh, Daniel Ziegler, Jeffrey Wu, Clemens Winter, Chris
  Hesse, Mark Chen, Eric Sigler, Mateusz Litwin, Scott Gray, Benjamin Chess,
  Jack Clark, Christopher Berner, Sam McCandlish, Alec Radford, Ilya Sutskever,
  and Dario Amodei. 2020.
\newblock \href
  {https://proceedings.neurips.cc/paper_files/paper/2020/file/1457c0d6bfcb4967418bfb8ac142f64a-Paper.pdf}
  {Language models are few-shot learners}.
\newblock In \emph{Advances in Neural Information Processing Systems},
  volume~33, pages 1877--1901. Curran Associates, Inc.

\bibitem[{Bulian et~al.(2022)Bulian, Buck, Gajewski, B{\"o}rschinger, and
  Schuster}]{bulian2022bem}
Jannis Bulian, Christian Buck, Wojciech Gajewski, Benjamin B{\"o}rschinger, and
  Tal Schuster. 2022.
\newblock \href {https://aclanthology.org/2022.emnlp-main.20} {Tomayto,
  tomahto. beyond token-level answer equivalence for question answering
  evaluation}.
\newblock In \emph{Proceedings of the 2022 Conference on Empirical Methods in
  Natural Language Processing}, pages 291--305, Abu Dhabi, United Arab
  Emirates. Association for Computational Linguistics.

\bibitem[{Chiang and Lee(2023)}]{chiang2023llmeval}
Cheng-Han Chiang and Hung-yi Lee. 2023.
\newblock \href {https://aclanthology.org/2023.acl-long.870} {Can large
  language models be an alternative to human evaluations?}
\newblock In \emph{Proceedings of the 61st Annual Meeting of the Association
  for Computational Linguistics (Volume 1: Long Papers)}, pages 15607--15631,
  Toronto, Canada. Association for Computational Linguistics.

\bibitem[{Chiang et~al.(2023)Chiang, Li, Lin, Sheng, Wu, Zhang, Zheng, Zhuang,
  Zhuang, Gonzalez, Stoica, and Xing}]{chiang2023vicuna}
Wei-Lin Chiang, Zhuohan Li, Zi~Lin, Ying Sheng, Zhanghao Wu, Hao Zhang, Lianmin
  Zheng, Siyuan Zhuang, Yonghao Zhuang, Joseph~E. Gonzalez, Ion Stoica, and
  Eric~P. Xing. 2023.
\newblock \href {https://lmsys.org/blog/2023-03-30-vicuna/} {Vicuna: An
  open-source chatbot impressing {GPT-4} with 90\%* {C}hat{GPT} quality}.

\bibitem[{Chiesurin et~al.(2023)Chiesurin, Dimakopoulos, Sobrevilla~Cabezudo,
  Eshghi, Papaioannou, Rieser, and Konstas}]{chiesurin2023dangers}
Sabrina Chiesurin, Dimitris Dimakopoulos, Marco~Antonio Sobrevilla~Cabezudo,
  Arash Eshghi, Ioannis Papaioannou, Verena Rieser, and Ioannis Konstas. 2023.
\newblock \href {https://aclanthology.org/2023.findings-acl.60} {The dangers of
  trusting stochastic parrots: Faithfulness and trust in open-domain
  conversational question answering}.
\newblock In \emph{Findings of the Association for Computational Linguistics:
  ACL 2023}, pages 947--959, Toronto, Canada. Association for Computational
  Linguistics.

\bibitem[{Christiano et~al.(2017)Christiano, Leike, Brown, Martic, Legg, and
  Amodei}]{christiano_deep_2017}
Paul~F Christiano, Jan Leike, Tom Brown, Miljan Martic, Shane Legg, and Dario
  Amodei. 2017.
\newblock \href
  {https://proceedings.neurips.cc/paper/2017/hash/d5e2c0adad503c91f91df240d0cd4e49-Abstract.html}
  {Deep {Reinforcement} {Learning} from {Human} {Preferences}}.
\newblock In \emph{Advances in {Neural} {Information} {Processing} {Systems}},
  volume~30. Curran Associates, Inc.

\bibitem[{Chung et~al.(2022)Chung, Hou, Longpre, Zoph, Tay, Fedus, Li, Wang,
  Dehghani, Brahma et~al.}]{chung2022flant5}
Hyung~Won Chung, Le~Hou, Shayne Longpre, Barret Zoph, Yi~Tay, William Fedus,
  Eric Li, Xuezhi Wang, Mostafa Dehghani, Siddhartha Brahma, et~al. 2022.
\newblock Scaling instruction-finetuned language models.
\newblock \emph{arXiv preprint arXiv:2210.11416}.

\bibitem[{Dziri et~al.(2022{\natexlab{a}})Dziri, Kamalloo, Milton, Zaiane, Yu,
  Ponti, and Reddy}]{dziri_faithdial_2022}
Nouha Dziri, Ehsan Kamalloo, Sivan Milton, Osmar Zaiane, Mo~Yu, Edoardo~M.
  Ponti, and Siva Reddy. 2022{\natexlab{a}}.
\newblock \href {https://doi.org/10.1162/tacl_a_00529} {Faithdial : {A}
  {Faithful} {Benchmark} for {Information}-{Seeking} {Dialogue}}.
\newblock \emph{Transactions of the Association for Computational Linguistics},
  10:1473--1490.

\bibitem[{Dziri et~al.(2023)Dziri, Lu, Sclar, Li, Jian, Lin, West, Bhagavatula,
  Bras, Hwang et~al.}]{dziri_faith_2023}
Nouha Dziri, Ximing Lu, Melanie Sclar, Xiang~Lorraine Li, Liwei Jian,
  Bill~Yuchen Lin, Peter West, Chandra Bhagavatula, Ronan~Le Bras, Jena~D
  Hwang, et~al. 2023.
\newblock Faith and fate: Limits of transformers on compositionality.
\newblock \emph{arXiv preprint arXiv:2305.18654}.

\bibitem[{Dziri et~al.(2022{\natexlab{b}})Dziri, Milton, Yu, Zaiane, and
  Reddy}]{dziri2022hallucination}
Nouha Dziri, Sivan Milton, Mo~Yu, Osmar Zaiane, and Siva Reddy.
  2022{\natexlab{b}}.
\newblock \href {https://doi.org/10.18653/v1/2022.naacl-main.387} {On the
  origin of hallucinations in conversational models: Is it the datasets or the
  models?}
\newblock In \emph{Proceedings of the 2022 Conference of the North American
  Chapter of the Association for Computational Linguistics: Human Language
  Technologies}, pages 5271--5285, Seattle, United States. Association for
  Computational Linguistics.

\bibitem[{Dziri et~al.(2022{\natexlab{c}})Dziri, Rashkin, Linzen, and
  Reitter}]{dziri-etal-2022-evaluating}
Nouha Dziri, Hannah Rashkin, Tal Linzen, and David Reitter. 2022{\natexlab{c}}.
\newblock \href {https://doi.org/10.1162/tacl_a_00506} {Evaluating attribution
  in dialogue systems: The {BEGIN} benchmark}.
\newblock \emph{Transactions of the Association for Computational Linguistics},
  10:1066--1083.

\bibitem[{Goyal et~al.(2022)Goyal, Li, and Durrett}]{goyal2022zeroshotnews}
Tanya Goyal, Junyi~Jessy Li, and Greg Durrett. 2022.
\newblock News summarization and evaluation in the era of {GPT-3}.
\newblock \emph{arXiv preprint arXiv:2209.12356}.

\bibitem[{Honovich et~al.(2021)Honovich, Choshen, Aharoni, Neeman, Szpektor,
  and Abend}]{honovich_q2_2021}
Or~Honovich, Leshem Choshen, Roee Aharoni, Ella Neeman, Idan Szpektor, and Omri
  Abend. 2021.
\newblock \href {https://doi.org/10.18653/v1/2021.emnlp-main.619}
  {Q$^{\textrm{2}}$: {Evaluating} {Factual} {Consistency} in
  {Knowledge}-{Grounded} {Dialogues} via {Question} {Generation} and {Question}
  {Answering}}.
\newblock In \emph{Proceedings of the 2021 {Conference} on {Empirical}
  {Methods} in {Natural} {Language} {Processing}}, pages 7856--7870, Online and
  Punta Cana, Dominican Republic. Association for Computational Linguistics.

\bibitem[{Iyer et~al.(2022)Iyer, Lin, Pasunuru, Mihaylov, Simig, Yu, Shuster,
  Wang, Liu, Koura et~al.}]{iyer2023optiml}
Srinivasan Iyer, Xi~Victoria Lin, Ramakanth Pasunuru, Todor Mihaylov, Daniel
  Simig, Ping Yu, Kurt Shuster, Tianlu Wang, Qing Liu, Punit~Singh Koura,
  et~al. 2022.
\newblock {OPT-IML}: Scaling language model instruction meta learning through
  the lens of generalization.
\newblock \emph{arXiv preprint arXiv:2212.12017}.

\bibitem[{Izacard and Grave(2021)}]{izacard2021fid}
Gautier Izacard and Edouard Grave. 2021.
\newblock \href {https://doi.org/10.18653/v1/2021.eacl-main.74} {Leveraging
  passage retrieval with generative models for open domain question answering}.
\newblock In \emph{Proceedings of the 16th Conference of the European Chapter
  of the Association for Computational Linguistics: Main Volume}, pages
  874--880, Online. Association for Computational Linguistics.

\bibitem[{Kamalloo et~al.(2023)Kamalloo, Dziri, Clarke, and
  Rafiei}]{kamalloo2023evaluating}
Ehsan Kamalloo, Nouha Dziri, Charles Clarke, and Davood Rafiei. 2023.
\newblock \href {https://aclanthology.org/2023.acl-long.307} {Evaluating
  open-domain question answering in the era of large language models}.
\newblock In \emph{Proceedings of the 61st Annual Meeting of the Association
  for Computational Linguistics (Volume 1: Long Papers)}, pages 5591--5606,
  Toronto, Canada. Association for Computational Linguistics.

\bibitem[{Karpukhin et~al.(2020)Karpukhin, Oguz, Min, Lewis, Wu, Edunov, Chen,
  and Yih}]{karpukhin2020dense}
Vladimir Karpukhin, Barlas Oguz, Sewon Min, Patrick Lewis, Ledell Wu, Sergey
  Edunov, Danqi Chen, and Wen-tau Yih. 2020.
\newblock \href {https://doi.org/10.18653/v1/2020.emnlp-main.550} {Dense
  passage retrieval for open-domain question answering}.
\newblock In \emph{Proceedings of the 2020 Conference on Empirical Methods in
  Natural Language Processing (EMNLP)}, pages 6769--6781, Online. Association
  for Computational Linguistics.

\bibitem[{Kendall(1945)}]{kendall-tau-b}
M.~G. Kendall. 1945.
\newblock \href {http://www.jstor.org/stable/2332303} {The treatment of ties in
  ranking problems}.
\newblock \emph{Biometrika}, 33(3):239--251.

\bibitem[{Kwiatkowski et~al.(2019)Kwiatkowski, Palomaki, Redfield, Collins,
  Parikh, Alberti, Epstein, Polosukhin, Devlin, Lee, Toutanova, Jones, Kelcey,
  Chang, Dai, Uszkoreit, Le, and Petrov}]{kwiatkowski2019nq}
Tom Kwiatkowski, Jennimaria Palomaki, Olivia Redfield, Michael Collins, Ankur
  Parikh, Chris Alberti, Danielle Epstein, Illia Polosukhin, Jacob Devlin,
  Kenton Lee, Kristina Toutanova, Llion Jones, Matthew Kelcey, Ming-Wei Chang,
  Andrew~M. Dai, Jakob Uszkoreit, Quoc Le, and Slav Petrov. 2019.
\newblock \href {https://doi.org/10.1162/tacl_a_00276} {Natural questions: A
  benchmark for question answering research}.
\newblock \emph{Transactions of the Association for Computational Linguistics},
  7:452--466.

\bibitem[{Lazaridou et~al.(2022)Lazaridou, Gribovskaya, Stokowiec, and
  Grigorev}]{lazaridou2022internetaugmentedlms}
Angeliki Lazaridou, Elena Gribovskaya, Wojciech Stokowiec, and Nikolai
  Grigorev. 2022.
\newblock Internet-augmented language models through few-shot prompting for
  open-domain question answering.
\newblock \emph{arXiv preprint arXiv:2203.05115}.

\bibitem[{Lee et~al.(2019)Lee, Chang, and Toutanova}]{lee2019orqa}
Kenton Lee, Ming-Wei Chang, and Kristina Toutanova. 2019.
\newblock \href {https://doi.org/10.18653/v1/P19-1612} {Latent retrieval for
  weakly supervised open domain question answering}.
\newblock In \emph{Proceedings of the 57th Annual Meeting of the Association
  for Computational Linguistics}, pages 6086--6096, Florence, Italy.

\bibitem[{Lewis et~al.(2020)Lewis, Perez, Piktus, Petroni, Karpukhin, Goyal,
  K\"{u}ttler, Lewis, Yih, Rockt\"{a}schel, Riedel, and Kiela}]{lewis2020rag}
Patrick Lewis, Ethan Perez, Aleksandra Piktus, Fabio Petroni, Vladimir
  Karpukhin, Naman Goyal, Heinrich K\"{u}ttler, Mike Lewis, Wen-tau Yih, Tim
  Rockt\"{a}schel, Sebastian Riedel, and Douwe Kiela. 2020.
\newblock \href
  {https://proceedings.neurips.cc/paper_files/paper/2020/file/6b493230205f780e1bc26945df7481e5-Paper.pdf}
  {Retrieval-augmented generation for knowledge-intensive {NLP} tasks}.
\newblock In \emph{Advances in Neural Information Processing Systems},
  volume~33, pages 9459--9474. Curran Associates, Inc.

\bibitem[{Lin(2004)}]{lin2004rouge}
Chin-Yew Lin. 2004.
\newblock \href {https://www.aclweb.org/anthology/W04-1013} {{ROUGE}: A package
  for automatic evaluation of summaries}.
\newblock In \emph{Text Summarization Branches Out}, pages 74--81, Barcelona,
  Spain. Association for Computational Linguistics.

\bibitem[{Liu et~al.(2023{\natexlab{a}})Liu, Lin, Hewitt, Paranjape,
  Bevilacqua, Petroni, and Liang}]{Liu023lostinmiddle}
Nelson~F Liu, Kevin Lin, John Hewitt, Ashwin Paranjape, Michele Bevilacqua,
  Fabio Petroni, and Percy Liang. 2023{\natexlab{a}}.
\newblock Lost in the middle: How language models use long contexts.
\newblock \emph{arXiv preprint arXiv:2307.03172}.

\bibitem[{Liu et~al.(2023{\natexlab{b}})Liu, Iter, Xu, Wang, Xu, and
  Zhu}]{liu2023geval}
Yang Liu, Dan Iter, Yichong Xu, Shuohang Wang, Ruochen Xu, and Chenguang Zhu.
  2023{\natexlab{b}}.
\newblock \href {https://doi.org/10.18653/v1/2023.emnlp-main.153} {{G}-eval:
  {NLG} evaluation using gpt-4 with better human alignment}.
\newblock In \emph{Proceedings of the 2023 Conference on Empirical Methods in
  Natural Language Processing}, pages 2511--2522, Singapore. Association for
  Computational Linguistics.

\bibitem[{Mallen et~al.(2023)Mallen, Asai, Zhong, Das, Khashabi, and
  Hajishirzi}]{Mallen2023llm_memorization}
Alex Mallen, Akari Asai, Victor Zhong, Rajarshi Das, Daniel Khashabi, and
  Hannaneh Hajishirzi. 2023.
\newblock \href {https://doi.org/10.18653/v1/2023.acl-long.546} {When not to
  trust language models: Investigating effectiveness of parametric and
  non-parametric memories}.
\newblock In \emph{Proceedings of the 61st Annual Meeting of the Association
  for Computational Linguistics (Volume 1: Long Papers)}, pages 9802--9822,
  Toronto, Canada. Association for Computational Linguistics.

\bibitem[{Min et~al.(2021)Min, Boyd-Graber, Alberti, Chen, Choi, Collins, Guu,
  Hajishirzi, Lee, Palomaki, Raffel, Roberts, Kwiatkowski, Lewis, Wu,
  K\"uttler, Liu, Minervini, Stenetorp, Riedel, Yang, Seo, Izacard, Petroni,
  Hosseini, Cao, Grave, Yamada, Shimaoka, Suzuki, Miyawaki, Sato, Takahashi,
  Suzuki, Fajcik, Docekal, Ondrej, Smrz, Cheng, Shen, Liu, He, Chen, Gao, Oguz,
  Chen, Karpukhin, Peshterliev, Okhonko, Schlichtkrull, Gupta, Mehdad, and
  Yih}]{min2021efficientqa}
Sewon Min, Jordan Boyd-Graber, Chris Alberti, Danqi Chen, Eunsol Choi, Michael
  Collins, Kelvin Guu, Hannaneh Hajishirzi, Kenton Lee, Jennimaria Palomaki,
  Colin Raffel, Adam Roberts, Tom Kwiatkowski, Patrick Lewis, Yuxiang Wu,
  Heinrich K\"uttler, Linqing Liu, Pasquale Minervini, Pontus Stenetorp,
  Sebastian Riedel, Sohee Yang, Minjoon Seo, Gautier Izacard, Fabio Petroni,
  Lucas Hosseini, Nicola~De Cao, Edouard Grave, Ikuya Yamada, Sonse Shimaoka,
  Masatoshi Suzuki, Shumpei Miyawaki, Shun Sato, Ryo Takahashi, Jun Suzuki,
  Martin Fajcik, Martin Docekal, Karel Ondrej, Pavel Smrz, Hao Cheng, Yelong
  Shen, Xiaodong Liu, Pengcheng He, Weizhu Chen, Jianfeng Gao, Barlas Oguz,
  Xilun Chen, Vladimir Karpukhin, Stan Peshterliev, Dmytro Okhonko, Michael
  Schlichtkrull, Sonal Gupta, Yashar Mehdad, and Wen-tau Yih. 2021.
\newblock \href {https://proceedings.mlr.press/v133/min21a.html} {Neur{IPS}
  2020 {EfficientQA} {Competition}: Systems, analyses and lessons learned}.
\newblock In \emph{Proceedings of the NeurIPS 2020 Competition and
  Demonstration Track}, volume 133, pages 86--111. Proceedings of Machine
  Learning Research.

\bibitem[{Mishra et~al.(2022)Mishra, Khashabi, Baral, and
  Hajishirzi}]{mishra2022naturalinstructions}
Swaroop Mishra, Daniel Khashabi, Chitta Baral, and Hannaneh Hajishirzi. 2022.
\newblock \href {https://doi.org/10.18653/v1/2022.acl-long.244} {Cross-task
  generalization via natural language crowdsourcing instructions}.
\newblock In \emph{Proceedings of the 60th Annual Meeting of the Association
  for Computational Linguistics (Volume 1: Long Papers)}, pages 3470--3487,
  Dublin, Ireland. Association for Computational Linguistics.

\bibitem[{OpenAI(2023)}]{openai2023gpt4}
OpenAI. 2023.
\newblock Gpt-4 technical report.
\newblock \emph{arXiv preprint arXiv:2303.08774}.

\bibitem[{Ouyang et~al.(2022)Ouyang, Wu, Jiang, Almeida, Wainwright, Mishkin,
  Zhang, Agarwal, Slama, Ray, Schulman, Hilton, Kelton, Miller, Simens, Askell,
  Welinder, Christiano, Leike, and Lowe}]{ouyang2022instructgpt}
Long Ouyang, Jeffrey Wu, Xu~Jiang, Diogo Almeida, Carroll Wainwright, Pamela
  Mishkin, Chong Zhang, Sandhini Agarwal, Katarina Slama, Alex Ray, John
  Schulman, Jacob Hilton, Fraser Kelton, Luke Miller, Maddie Simens, Amanda
  Askell, Peter Welinder, Paul~F Christiano, Jan Leike, and Ryan Lowe. 2022.
\newblock \href
  {https://proceedings.neurips.cc/paper_files/paper/2022/file/b1efde53be364a73914f58805a001731-Paper-Conference.pdf}
  {Training language models to follow instructions with human feedback}.
\newblock In \emph{Advances in Neural Information Processing Systems},
  volume~35, pages 27730--27744. Curran Associates, Inc.

\bibitem[{Peng et~al.(2023)Peng, Li, He, Galley, and Gao}]{peng2023instruction}
Baolin Peng, Chunyuan Li, Pengcheng He, Michel Galley, and Jianfeng Gao. 2023.
\newblock Instruction tuning with gpt-4.
\newblock \emph{arXiv preprint arXiv:2304.03277}.

\bibitem[{Raffel et~al.(2020)Raffel, Shazeer, Roberts, Lee, Narang, Matena,
  Zhou, Li, and Liu}]{raffel2020t5}
Colin Raffel, Noam Shazeer, Adam Roberts, Katherine Lee, Sharan Narang, Michael
  Matena, Yanqi Zhou, Wei Li, and Peter~J. Liu. 2020.
\newblock \href {http://jmlr.org/papers/v21/20-074.html} {Exploring the limits
  of transfer learning with a unified text-to-text transformer}.
\newblock \emph{Journal of Machine Learning Research}, 21(140):1--67.

\bibitem[{Rajpurkar et~al.(2016)Rajpurkar, Zhang, Lopyrev, and
  Liang}]{rajpurkar-etal-2016-squad}
Pranav Rajpurkar, Jian Zhang, Konstantin Lopyrev, and Percy Liang. 2016.
\newblock \href {https://doi.org/10.18653/v1/D16-1264} {{SQ}u{AD}: 100,000+
  questions for machine comprehension of text}.
\newblock In \emph{Proceedings of the 2016 Conference on Empirical Methods in
  Natural Language Processing}, pages 2383--2392, Austin, Texas. Association
  for Computational Linguistics.

\bibitem[{Rashkin et~al.(2021{\natexlab{a}})Rashkin, Nikolaev, Lamm, Collins,
  Das, Petrov, Tomar, Turc, and Reitter}]{Rashkin2021attribution}
Hannah Rashkin, Vitaly Nikolaev, Matthew Lamm, Michael Collins, Dipanjan Das,
  Slav Petrov, Gaurav~Singh Tomar, Iulia Turc, and D.~Reitter.
  2021{\natexlab{a}}.
\newblock Measuring attribution in natural language generation models.
\newblock \emph{ArXiv}, abs/2112.12870.

\bibitem[{Rashkin et~al.(2021{\natexlab{b}})Rashkin, Reitter, Tomar, and
  Das}]{rashkin-etal-2021-increasing}
Hannah Rashkin, David Reitter, Gaurav~Singh Tomar, and Dipanjan Das.
  2021{\natexlab{b}}.
\newblock \href {https://doi.org/10.18653/v1/2021.acl-long.58} {Increasing
  faithfulness in knowledge-grounded dialogue with controllable features}.
\newblock In \emph{Proceedings of the 59th Annual Meeting of the Association
  for Computational Linguistics and the 11th International Joint Conference on
  Natural Language Processing (Volume 1: Long Papers)}, pages 704--718, Online.
  Association for Computational Linguistics.

\bibitem[{Reddy et~al.(2019)Reddy, Chen, and Manning}]{reddy-etal-2019-coqa}
Siva Reddy, Danqi Chen, and Christopher~D. Manning. 2019.
\newblock \href {https://doi.org/10.1162/tacl_a_00266} {{C}o{QA}: A
  conversational question answering challenge}.
\newblock \emph{Transactions of the Association for Computational Linguistics},
  7:249--266.

\bibitem[{Robertson et~al.(1995)Robertson, Walker, Jones, Hancock-Beaulieu,
  Gatford et~al.}]{robertson1995okapi}
Stephen~E Robertson, Steve Walker, Susan Jones, Micheline~M Hancock-Beaulieu,
  Mike Gatford, et~al. 1995.
\newblock Okapi at trec-3.
\newblock \emph{Nist Special Publication Sp}, 109:109.

\bibitem[{Sanh et~al.(2022)Sanh, Webson, Raffel, Bach, Sutawika, Alyafeai,
  Chaffin, Stiegler, Raja, Dey, Bari, Xu, Thakker, Sharma, Szczechla, Kim,
  Chhablani, Nayak, Datta, Chang, Jiang, Wang, Manica, Shen, Yong, Pandey,
  Bawden, Wang, Neeraj, Rozen, Sharma, Santilli, Fevry, Fries, Teehan, Scao,
  Biderman, Gao, Wolf, and Rush}]{sanh2022t0}
Victor Sanh, Albert Webson, Colin Raffel, Stephen Bach, Lintang Sutawika, Zaid
  Alyafeai, Antoine Chaffin, Arnaud Stiegler, Arun Raja, Manan Dey, M~Saiful
  Bari, Canwen Xu, Urmish Thakker, Shanya~Sharma Sharma, Eliza Szczechla,
  Taewoon Kim, Gunjan Chhablani, Nihal Nayak, Debajyoti Datta, Jonathan Chang,
  Mike Tian-Jian Jiang, Han Wang, Matteo Manica, Sheng Shen, Zheng~Xin Yong,
  Harshit Pandey, Rachel Bawden, Thomas Wang, Trishala Neeraj, Jos Rozen,
  Abheesht Sharma, Andrea Santilli, Thibault Fevry, Jason~Alan Fries, Ryan
  Teehan, Teven~Le Scao, Stella Biderman, Leo Gao, Thomas Wolf, and Alexander~M
  Rush. 2022.
\newblock \href {https://openreview.net/forum?id=9Vrb9D0WI4} {Multitask
  prompted training enables zero-shot task generalization}.
\newblock In \emph{International Conference on Learning Representations}.

\bibitem[{Shi et~al.(2023)Shi, Min, Yasunaga, Seo, James, Lewis, Zettlemoyer,
  and tau Yih}]{shi2023replug}
Weijia Shi, Sewon Min, Michihiro Yasunaga, Minjoon Seo, Rich James, Mike Lewis,
  Luke Zettlemoyer, and Wen tau Yih. 2023.
\newblock \href {http://arxiv.org/abs/2301.12652} {{REPLUG}:
  Retrieval-augmented black-box language models}.

\bibitem[{Shuster et~al.(2021)Shuster, Poff, Chen, Kiela, and
  Weston}]{shuster2021retrieval}
Kurt Shuster, Spencer Poff, Moya Chen, Douwe Kiela, and Jason Weston. 2021.
\newblock \href {https://doi.org/10.18653/v1/2021.findings-emnlp.320}
  {Retrieval augmentation reduces hallucination in conversation}.
\newblock In \emph{Findings of the Association for Computational Linguistics:
  EMNLP 2021}, pages 3784--3803, Punta Cana, Dominican Republic. Association
  for Computational Linguistics.

\bibitem[{Sidiropoulos et~al.(2021)Sidiropoulos, Voskarides, Vakulenko, and
  Kanoulas}]{sidiropoulos-etal-2021-combining}
Georgios Sidiropoulos, Nikos Voskarides, Svitlana Vakulenko, and Evangelos
  Kanoulas. 2021.
\newblock \href {https://doi.org/10.18653/v1/2021.sustainlp-1.7} {Combining
  lexical and dense retrieval for computationally efficient multi-hop question
  answering}.
\newblock In \emph{Proceedings of the Second Workshop on Simple and Efficient
  Natural Language Processing}, pages 58--63, Virtual. Association for
  Computational Linguistics.

\bibitem[{Taori et~al.(2023)Taori, Gulrajani, Zhang, Dubois, Li, Guestrin,
  Liang, and Hashimoto}]{Taori2023alpaca}
Rohan Taori, Ishaan Gulrajani, Tianyi Zhang, Yann Dubois, Xuechen Li, Carlos
  Guestrin, Percy Liang, and Tatsunori~B. Hashimoto. 2023.
\newblock {Stanford Alpaca}: An instruction-following {LLaMA} model.
\newblock \url{https://github.com/tatsu-lab/stanford_alpaca}.

\bibitem[{Touvron et~al.(2023{\natexlab{a}})Touvron, Lavril, Izacard, Martinet,
  Lachaux, Lacroix, Rozière, Goyal, Hambro, Azhar, Rodriguez, Joulin, Grave,
  and Lample}]{touvron2023llama}
Hugo Touvron, Thibaut Lavril, Gautier Izacard, Xavier Martinet, Marie-Anne
  Lachaux, Timothée Lacroix, Baptiste Rozière, Naman Goyal, Eric Hambro,
  Faisal Azhar, Aurelien Rodriguez, Armand Joulin, Edouard Grave, and Guillaume
  Lample. 2023{\natexlab{a}}.
\newblock \href {http://arxiv.org/abs/2302.13971} {{LLaMA}: Open and efficient
  foundation language models}.

\bibitem[{Touvron et~al.(2023{\natexlab{b}})Touvron, Martin, Stone, Albert,
  Almahairi, Babaei, Bashlykov, Batra, Bhargava, Bhosale, Bikel, Blecher,
  Ferrer, Chen, Cucurull, Esiobu, Fernandes, Fu, Fu, Fuller, Gao, Goswami,
  Goyal, Hartshorn, Hosseini, Hou, Inan, Kardas, Kerkez, Khabsa, Kloumann,
  Korenev, Koura, Lachaux, Lavril, Lee, Liskovich, Lu, Mao, Martinet, Mihaylov,
  Mishra, Molybog, Nie, Poulton, Reizenstein, Rungta, Saladi, Schelten, Silva,
  Smith, Subramanian, Tan, Tang, Taylor, Williams, Kuan, Xu, Yan, Zarov, Zhang,
  Fan, Kambadur, Narang, Rodriguez, Stojnic, Edunov, and
  Scialom}]{touvron2023llama2}
Hugo Touvron, Louis Martin, Kevin Stone, Peter Albert, Amjad Almahairi, Yasmine
  Babaei, Nikolay Bashlykov, Soumya Batra, Prajjwal Bhargava, Shruti Bhosale,
  Dan Bikel, Lukas Blecher, Cristian~Canton Ferrer, Moya Chen, Guillem
  Cucurull, David Esiobu, Jude Fernandes, Jeremy Fu, Wenyin Fu, Brian Fuller,
  Cynthia Gao, Vedanuj Goswami, Naman Goyal, Anthony Hartshorn, Saghar
  Hosseini, Rui Hou, Hakan Inan, Marcin Kardas, Viktor Kerkez, Madian Khabsa,
  Isabel Kloumann, Artem Korenev, Punit~Singh Koura, Marie-Anne Lachaux,
  Thibaut Lavril, Jenya Lee, Diana Liskovich, Yinghai Lu, Yuning Mao, Xavier
  Martinet, Todor Mihaylov, Pushkar Mishra, Igor Molybog, Yixin Nie, Andrew
  Poulton, Jeremy Reizenstein, Rashi Rungta, Kalyan Saladi, Alan Schelten, Ruan
  Silva, Eric~Michael Smith, Ranjan Subramanian, Xiaoqing~Ellen Tan, Binh Tang,
  Ross Taylor, Adina Williams, Jian~Xiang Kuan, Puxin Xu, Zheng Yan, Iliyan
  Zarov, Yuchen Zhang, Angela Fan, Melanie Kambadur, Sharan Narang, Aurelien
  Rodriguez, Robert Stojnic, Sergey Edunov, and Thomas Scialom.
  2023{\natexlab{b}}.
\newblock \href {http://arxiv.org/abs/2307.09288} {Llama 2: Open foundation and
  fine-tuned chat models}.

\bibitem[{Wang et~al.(2023)Wang, Li, Chen, Zhu, Lin, Cao, Liu, Liu, and
  Sui}]{wang2023llmfaireval}
Peiyi Wang, Lei Li, Liang Chen, Dawei Zhu, Binghuai Lin, Yunbo Cao, Qi~Liu,
  Tianyu Liu, and Zhifang Sui. 2023.
\newblock Large language models are not fair evaluators.
\newblock \emph{arXiv preprint arXiv:2305.17926}.

\bibitem[{Wang et~al.(2022{\natexlab{a}})Wang, Kordi, Mishra, Liu, Smith,
  Khashabi, and Hajishirzi}]{wang2022selfinstruct}
Yizhong Wang, Yeganeh Kordi, Swaroop Mishra, Alisa Liu, Noah~A. Smith, Daniel
  Khashabi, and Hannaneh Hajishirzi. 2022{\natexlab{a}}.
\newblock Self-instruct: Aligning language model with self generated
  instructions.
\newblock \emph{arXiv preprint arXiv:2212.10560}.

\bibitem[{Wang et~al.(2022{\natexlab{b}})Wang, Mishra, Alipoormolabashi, Kordi,
  Mirzaei, Naik, Ashok, Dhanasekaran, Arunkumar, Stap, Pathak, Karamanolakis,
  Lai, Purohit, Mondal, Anderson, Kuznia, Doshi, Pal, Patel, Moradshahi,
  Parmar, Purohit, Varshney, Kaza, Verma, Puri, Karia, Doshi, Sampat, Mishra,
  Reddy~A, Patro, Dixit, and Shen}]{wang2022super}
Yizhong Wang, Swaroop Mishra, Pegah Alipoormolabashi, Yeganeh Kordi, Amirreza
  Mirzaei, Atharva Naik, Arjun Ashok, Arut~Selvan Dhanasekaran, Anjana
  Arunkumar, David Stap, Eshaan Pathak, Giannis Karamanolakis, Haizhi Lai,
  Ishan Purohit, Ishani Mondal, Jacob Anderson, Kirby Kuznia, Krima Doshi,
  Kuntal~Kumar Pal, Maitreya Patel, Mehrad Moradshahi, Mihir Parmar, Mirali
  Purohit, Neeraj Varshney, Phani~Rohitha Kaza, Pulkit Verma, Ravsehaj~Singh
  Puri, Rushang Karia, Savan Doshi, Shailaja~Keyur Sampat, Siddhartha Mishra,
  Sujan Reddy~A, Sumanta Patro, Tanay Dixit, and Xudong Shen.
  2022{\natexlab{b}}.
\newblock \href {https://aclanthology.org/2022.emnlp-main.340}
  {Super-{N}atural{I}nstructions: Generalization via declarative instructions
  on 1600+ {NLP} tasks}.
\newblock In \emph{Proceedings of the 2022 Conference on Empirical Methods in
  Natural Language Processing}, pages 5085--5109, Abu Dhabi, United Arab
  Emirates. Association for Computational Linguistics.

\bibitem[{Wei et~al.(2022)Wei, Bosma, Zhao, Guu, Yu, Lester, Du, Dai, and
  Le}]{wei2022flan}
Jason Wei, Maarten Bosma, Vincent Zhao, Kelvin Guu, Adams~Wei Yu, Brian Lester,
  Nan Du, Andrew~M. Dai, and Quoc~V Le. 2022.
\newblock \href {https://openreview.net/forum?id=gEZrGCozdqR} {Finetuned
  language models are zero-shot learners}.
\newblock In \emph{International Conference on Learning Representations}.

\bibitem[{Xiong et~al.(2021)Xiong, Li, Iyer, Du, Lewis, Wang, Mehdad, Yih,
  Riedel, Kiela, and Oguz}]{xiong2021mhdr}
Wenhan Xiong, Xiang Li, Srini Iyer, Jingfei Du, Patrick Lewis, William~Yang
  Wang, Yashar Mehdad, Scott Yih, Sebastian Riedel, Douwe Kiela, and Barlas
  Oguz. 2021.
\newblock \href {https://openreview.net/forum?id=EMHoBG0avc1} {Answering
  complex open-domain questions with multi-hop dense retrieval}.
\newblock In \emph{International Conference on Learning Representations}.

\bibitem[{Xu et~al.(2023)Xu, Song, Iyyer, and Choi}]{xu2023lfqaevaluation}
Fangyuan Xu, Yixiao Song, Mohit Iyyer, and Eunsol Choi. 2023.
\newblock \href {https://doi.org/10.18653/v1/2023.acl-long.181} {A critical
  evaluation of evaluations for long-form question answering}.
\newblock In \emph{Proceedings of the 61st Annual Meeting of the Association
  for Computational Linguistics (Volume 1: Long Papers)}, pages 3225--3245,
  Toronto, Canada. Association for Computational Linguistics.

\bibitem[{Yang et~al.(2018)Yang, Qi, Zhang, Bengio, Cohen, Salakhutdinov, and
  Manning}]{yang2018hotpotqa}
Zhilin Yang, Peng Qi, Saizheng Zhang, Yoshua Bengio, William~W. Cohen, Ruslan
  Salakhutdinov, and Christopher~D. Manning. 2018.
\newblock {HotpotQA}: A dataset for diverse, explainable multi-hop question
  answering.
\newblock In \emph{Conference on Empirical Methods in Natural Language
  Processing ({EMNLP})}.

\bibitem[{Zhang et~al.(2022)Zhang, Roller, Goyal, Artetxe, Chen, Chen, Dewan,
  Diab, Li, Lin, Mihaylov, Ott, Shleifer, Shuster, Simig, Koura, Sridhar, Wang,
  and Zettlemoyer}]{zhang2022opt}
Susan Zhang, Stephen Roller, Naman Goyal, Mikel Artetxe, Moya Chen, Shuohui
  Chen, Christopher Dewan, Mona Diab, Xian Li, Xi~Victoria Lin, Todor Mihaylov,
  Myle Ott, Sam Shleifer, Kurt Shuster, Daniel Simig, Punit~Singh Koura, Anjali
  Sridhar, Tianlu Wang, and Luke Zettlemoyer. 2022.
\newblock \href {http://arxiv.org/abs/2205.01068} {{OPT}: Open pre-trained
  transformer language models}.

\bibitem[{Zhang et~al.(2020)Zhang, Kishore, Wu, Weinberger, and
  Artzi}]{zhang2020bertscore}
Tianyi Zhang, Varsha Kishore, Felix Wu, Kilian~Q. Weinberger, and Yoav Artzi.
  2020.
\newblock \href {https://openreview.net/forum?id=SkeHuCVFDr} {{BERTScore}:
  Evaluating text generation with {BERT}}.
\newblock In \emph{International Conference on Learning Representations}.

\end{thebibliography}
